\documentclass[lettersize,journal,onecolum]{IEEEtran}
\usepackage{diagbox}
\usepackage{algorithmicx,algorithm}
\usepackage{graphicx}
\usepackage{amsmath}
\usepackage{amssymb}
\usepackage{booktabs}
\usepackage{ulem}
\graphicspath{{./Figs/}}
\newcommand{\argmin}{\operatornamewithlimits{argmin}}
\usepackage[pagebackref,breaklinks,colorlinks]{hyperref}
\newcommand{\etal}{\textit{et al. }}
\usepackage[capitalize]{cleveref}
\crefname{section}{Sec.}{Secs.}
\Crefname{section}{Section}{Sections}
\Crefname{table}{Table}{Tables}
\crefname{table}{Tab.}{Tabs.}
\usepackage[caption=false,font=normalsize,labelfont=sf,textfont=sf]{subfig}

\usepackage[section]{placeins}

\begin{document}

\title{SDIP: Self-Reinforcement Deep Image Prior Framework for Image Processing}

\author{Ziyu Shu and Zhixin Pan
\thanks{Ziyu Shu (corresponding author) is with the Washington University in St Louis, 63130, USA (email: zs919@nyu.edu).}
\thanks{Zhixin Pan is with the Florida State University, 32306, USA (email:zp23e@fsu.edu).}}

\maketitle

\begin{abstract}
Deep image prior (DIP) proposed in recent research has revealed the inherent trait of convolutional neural networks (CNN) for capturing substantial low-level image statistics priors. This framework efficiently addresses the inverse problems in image processing and has induced extensive applications in various domains. However, as the whole algorithm is initialized randomly, the DIP algorithm often lacks stability. Thus, this method still has space for further improvement. In this paper, we propose the self-reinforcement deep image prior (SDIP) as an improved version of the original DIP. We observed that the changes in the DIP networks' input and output are highly correlated during each iteration. SDIP efficiently utilizes this trait in a reinforcement learning manner, where the current iteration's output is utilized by a steering algorithm to update the network input for the next iteration, guiding the algorithm toward improved results. Experimental results across multiple applications demonstrate that our proposed SDIP framework offers improvement compared to the original DIP method and other state-of-the-art methods.

\end{abstract}


\section{Introduction}
\label{sec:intro}

Image recovery is now extensively applied in various fields, including but not limited to medical image reconstruction~\cite{willemink2019evolution}, deblurring~\cite{li2022survey}, and super-resolution~\cite{lepcha2023image}. This type of task is commonly framed as an inverse problem, aiming to recover the original image $\boldsymbol x$ based on available measurement $\boldsymbol y$. Currently, researchers mainly use linear optimization algorithms to solve these problems, they leverage mathematical models for acquisition as well as image priors into an optimization process. High-quality results can be obtained by minimizing an objective function~\cite{shewchuk1994introduction}. However, linear optimization algorithms often struggle with highly ill-posed inverse problems~\cite{engl2014inverse,kabanikhin2008definitions}, which are prevalent in the tasks mentioned above.

To overcome this challenge, researchers such as Wang \etal proposed using pre-trained neural networks~\cite{wang2018image}. The superior performance of the pre-trained neural networks is imputed to their ability to learn extra information from training data. Thus, a well-trained neural network can be regarded as a powerful application-specific prior. While these methods have already made impressive achievements~\cite{jin2017deep}, their demand for a substantial amount of high-quality training data and their lack of stability limit their broader applicability~\cite{antun2020instabilities, bhadra2021hallucinations}.

The deep image prior (DIP)~\cite{ulyanov2018deep} introduced by Ulyanov \etal presents new possibilities for addressing highly ill-posed inverse problems. The author pointed out that the hierarchical structure of a convolutional neural network (CNN) itself inherently possesses the capability to capture abundant low-level image statistical prior, thereby enabling the generation of high-quality images for inverse problems without any training process. Consequently, a randomly initialized CNN can be used for inverse problems by optimizing its network weights to minimize the same objective functions used with linear optimization algorithms~\cite{dittmer2020regularization}.

While DIP has proven to be quite effective and demonstrated success in several inverse problems, its results still lag behind some of the unsupervised state-of-the-art alternatives. Multiple attempts have been made to improve its performance: Liu \etal suggested augmenting the loss function with extra regularization~\cite{liu2019image}, such as total variation. Cui \etal introduced the Conditional DIP, which utilizes a reference image similar to the ground truth image as the network's input to guide the DIP network~\cite{cui2019ct}. Mataev \etal advocated for the use of the ADMM (alternating direction method of multipliers) framework to integrate denoisers like the RED (Regularization by Denoising) algorithm into the DIP methods~\cite{mataev2019deepred}. More researches are available from \cite{jo2021rethinking,shu2022sparse, shu2022rbp, xu2023deep,li2023enhanced,cui2022unsupervised}

In this paper, we proposed the SDIP (self-reinforcement deep image prior), a novel method that can deliver high-quality results for multiple highly ill-posed problems. This newly proposed method enables the enhancement of the original DIP algorithm by leveraging most existing algorithms to assist it, thereby significantly improving its stability and exceptionally high adaptability to highly ill-posed problems. Furthermore, it requires no training data. 

Specifically, this paper makes the following contributions:
\begin{itemize}
\item SDIP leverages the inherent correlation between changes in network input and network output, combining deep image prior with priors integrated in classic linear optimization algorithms. As a result, it combines the advantages of both methods while mitigating their shortcomings.
\item To the best of our knowledge, our proposed approach, SDIP, is the first attempt of using the self-reinforcement mechanism to introduce extra priors to the DIP framework. This mechanism exhibits great scalability, as most priors and related methods can also be incorporated to further improve the algorithm's performance. 
\item Experimental evaluation demonstrates that SDIP exhibits a versatile enhancement of the original DIP algorithm across multiple applications. 
\end{itemize}

The remainder of this paper is organized as follows: the details of the interested inverse problem and the deep image prior property are introduced in Section \ref{sec:related}. The proposed SDIP framework is introduced in Section \ref{sec:proposedmethod}, and compared with the original DIP framework in Section \ref{sec:app}. The results from our experiments are discussed in Section \ref{sec:discussion}, followed by the future work Section \ref{sec:future} and the conclusion Section \ref{sec:conclusions}.

\section{Related Works}
\label{sec:related}
In this section, we first provide the background of \textit{inverse problem} which can universally formulate most image recovery tasks we are addressing. Next, we introduce the \textit{deep image prior} (DIP) as a novel approach for solving inverse problems without the need for a training process or training dataset.
\subsection{The Inverse Problems }
Formally, an inverse problem of our research interest focuses on cases where the measurement $\boldsymbol y$ is defined as $\boldsymbol y= \boldsymbol{\rm H} \boldsymbol x+ \boldsymbol \epsilon$, with $\boldsymbol \epsilon$ representing noises. Through adjustments to the operator $\boldsymbol{\rm H}$, we have the flexibility to switch between various image recovering problems, including denoising ($\boldsymbol{\rm H} = \boldsymbol{\rm I}$), deblurring ($\boldsymbol{\rm H}$ is a convolution filter), super-resolution ($\boldsymbol{\rm H}$ represents both a blur followed by a sub-sampling), and tomographic reconstruction ($\boldsymbol{\rm H}$ is the X-ray transform).

Conventional linear optimization methods proposed solving these problems by minimizing the following objective function:
\begin{equation}
\boldsymbol x^* = \argmin_{x}  ||\boldsymbol y - \boldsymbol{\rm H} \boldsymbol x||^2_2 + \lambda f(\boldsymbol x),
\label{linearop}
\end{equation}
where $f(\boldsymbol x)$ serves as a regularization term. In cases where the inverse problem is well-posed, having a unique solution to minimize the data inconsistency term $||\boldsymbol y - \boldsymbol{\rm H} \boldsymbol x||^2_2$, regularization primarily addresses the balance between data fidelity and image regularity for denoising purposes. Conversely, in ill-posed problems where multiple solutions exist for minimizing the data fidelity term, regularization serves as a crucial prior, aiding in the selection of the most plausible solution among all candidates. However, in highly ill-posed problems, achieving high-quality results can be challenging, even without measurement noise, when utilizing manually crafted priors such as sparsity-promoting regularizations.

To this end, researchers proposed using deep learning approaches that are trained on large datasets, as a well-trained neural network can be regarded as a potent application-specific prior. Despite numerous experiments and applications demonstrating its effectiveness in addressing the aforementioned inverse problems, its utility remains constrained by the substantial need for large quantities of high-quality training data and inherent instability issues~\cite{antun2020instabilities, bhadra2021hallucinations}.

\subsection{Deep Image Prior}
To overcome these challenges, \textit{Deep Image Prior} (DIP) proposed in~\cite{ulyanov2018deep} takes a different route without the need for any training data or training process. The author pointed out that the structure of a convolutional neural network itself is a powerful prior for generating general images. In other words, the results of these inverse problems lie in the space spanned by an untrained convolutional neural network. Consequently, a randomly initialized convolutional neural network can be used to solve the aforementioned inverse problem, which can be expressed as:
\begin{equation}
\begin{aligned}
\boldsymbol \theta^* &= \argmin_{\theta}  ||\boldsymbol y - \boldsymbol{\rm H} G(\boldsymbol \theta |\boldsymbol z)||^2_2,\\
\boldsymbol x^* &= G(\boldsymbol \theta^* |\boldsymbol z),
\end{aligned}
\label{dipop}
\end{equation}
where $G$ represents a convolutional neural network. Its weights $\boldsymbol \theta$ are randomly initialized and will be optimized during the optimization process, and the network input $\boldsymbol z$ is a randomly initialized vector that will be fixed during the optimization process. A high-quality result $\boldsymbol x^*$ can be obtained once the objective function is minimized.

Compared to the conventional linear optimization methods, although both the methods optimize a similar objective function (most regularization term $f(\boldsymbol x)$ in Equation \ref{linearop} can also be added into Equation \ref{dipop} in the form as $f(G(\boldsymbol \theta |\boldsymbol z))$~\cite{liu2019image}). The use of the untrained convolutional neural network $G$ utilizes the DIP and thus has the potential to achieve better results. Compared to pre-trained methods, DIP related methods eliminate the interference from the training process. Furthermore, neural networks in DIP related methods minimize objective functions on the inference data but not training data, so that at least a local minimum can be found. This also implies that the result generated by DIP methods is at least not worse than that of linear optimization methods in terms of the objective functions if both attain a similar loss. Recently, various algorithms based on DIP have been proposed and have achieved impressive results~\cite{shu2022sparse,shu2022rbp,jo2021rethinking,heckel2019denoising,zhao2020reference}. However, these algorithms are often far from mature. This is primarily because the DIP network, which forms the core of these algorithms, is randomly generated. This uncertainty makes the results of the algorithm unstable and often prevents it from achieving the optimum.

\section{Proposed Method}
\label{sec:proposedmethod}
In this section, we explained our proposed Self-Reinforcement Deep Image Prior (SDIP) framework in detail. First, we introduce the motivation, where two key observations are demonstrated: i) There is a close correlation between initial input and final output from DIP, and ii) modifying network input produces statistically identical effects on network output during DIP optimization. Then, we combine the above two observations and provide a step-by-step explanation of the proposed self-reinforcement deep image prior method.

\subsection{Correlation Between the Initial Input and Final Output in DIP}
The success of CDIP implies that the input to DIP networks can significantly affect the convergence speed and the quality of the final results. We provide the following examples to illustrate this property.

For the sake of simplicity, the operator $\boldsymbol{\rm H}$ in the following image recovering task is set to the identity matrix $\boldsymbol{\rm I}$, and noise is not considered. In that case, the problem can be expressed as:
\begin{equation}
\begin{aligned}
\boldsymbol \theta^* &= \argmin_{\theta}  ||\boldsymbol x - G(\boldsymbol \theta |\boldsymbol z)||^2_2,\\
\boldsymbol x^* &= G(\boldsymbol \theta^* |\boldsymbol z),
\end{aligned}
\label{cdipexp}
\end{equation}
where the structure of network $G$ is the standard U-net, our objective is to investigate the accuracy and convergence speed of the algorithm when using different network input $\boldsymbol z$.

\begin{figure}[htpb]
  \centering
   \includegraphics[width=1\linewidth]{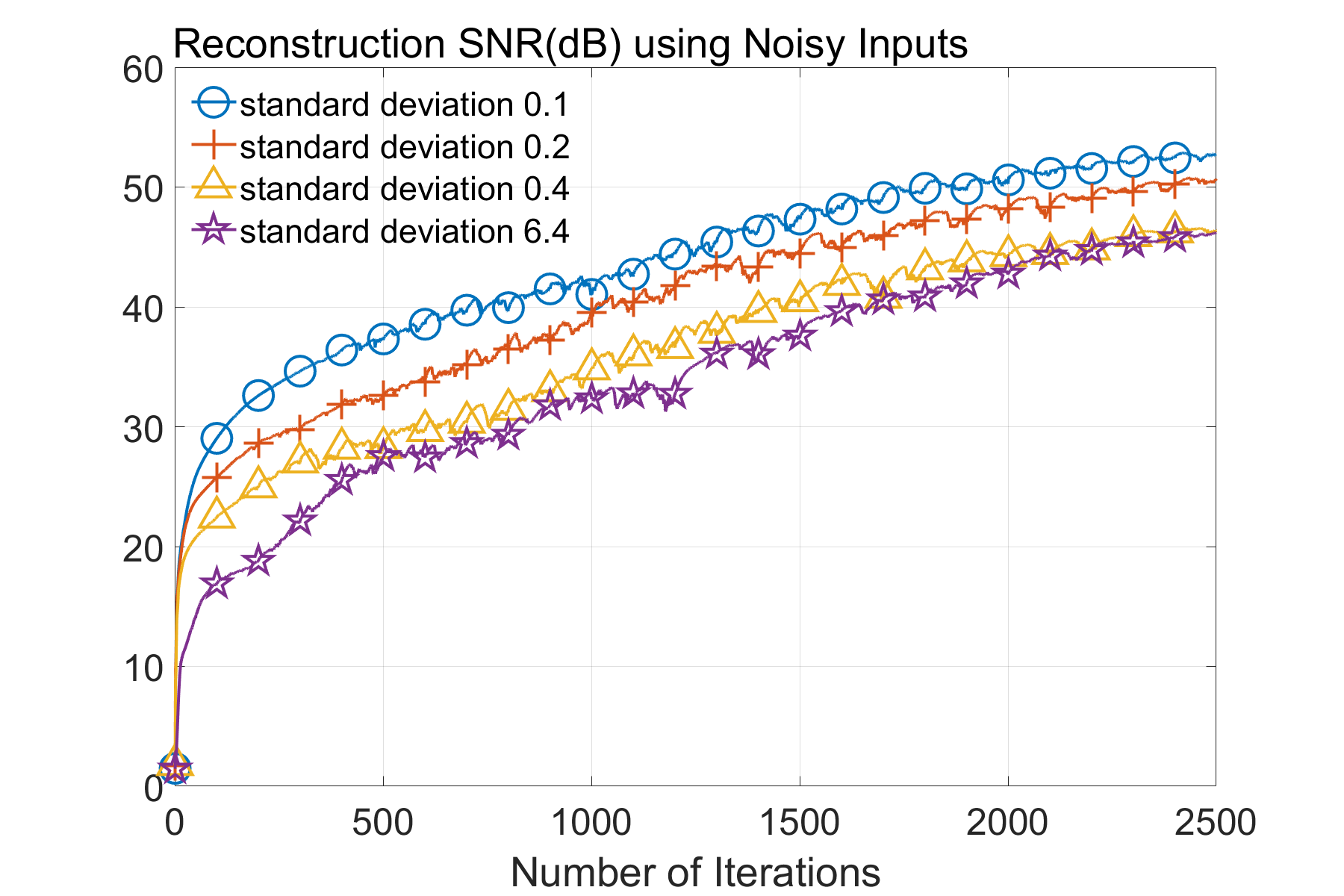}
   \caption{The reconstruction SNR under different numbers of iterations with noisy inputs.}
   \label{fig:gaussian inputs}
\end{figure}

In our first example, the network input is configured as $\boldsymbol z(i,j) = \mathcal{N}(\boldsymbol x(i,j), \sigma)$ with varying standard deviations. So $\boldsymbol z$ approaches to $\boldsymbol x$ as $\sigma$ tends to $0$. The corresponding result is shown in Fig.\ref{fig:gaussian inputs}, clearly showing the strong correlation between the network's input and reconstruction accuracy. 

\begin{figure}[htpb]
  \centering
   \includegraphics[width=1\linewidth]{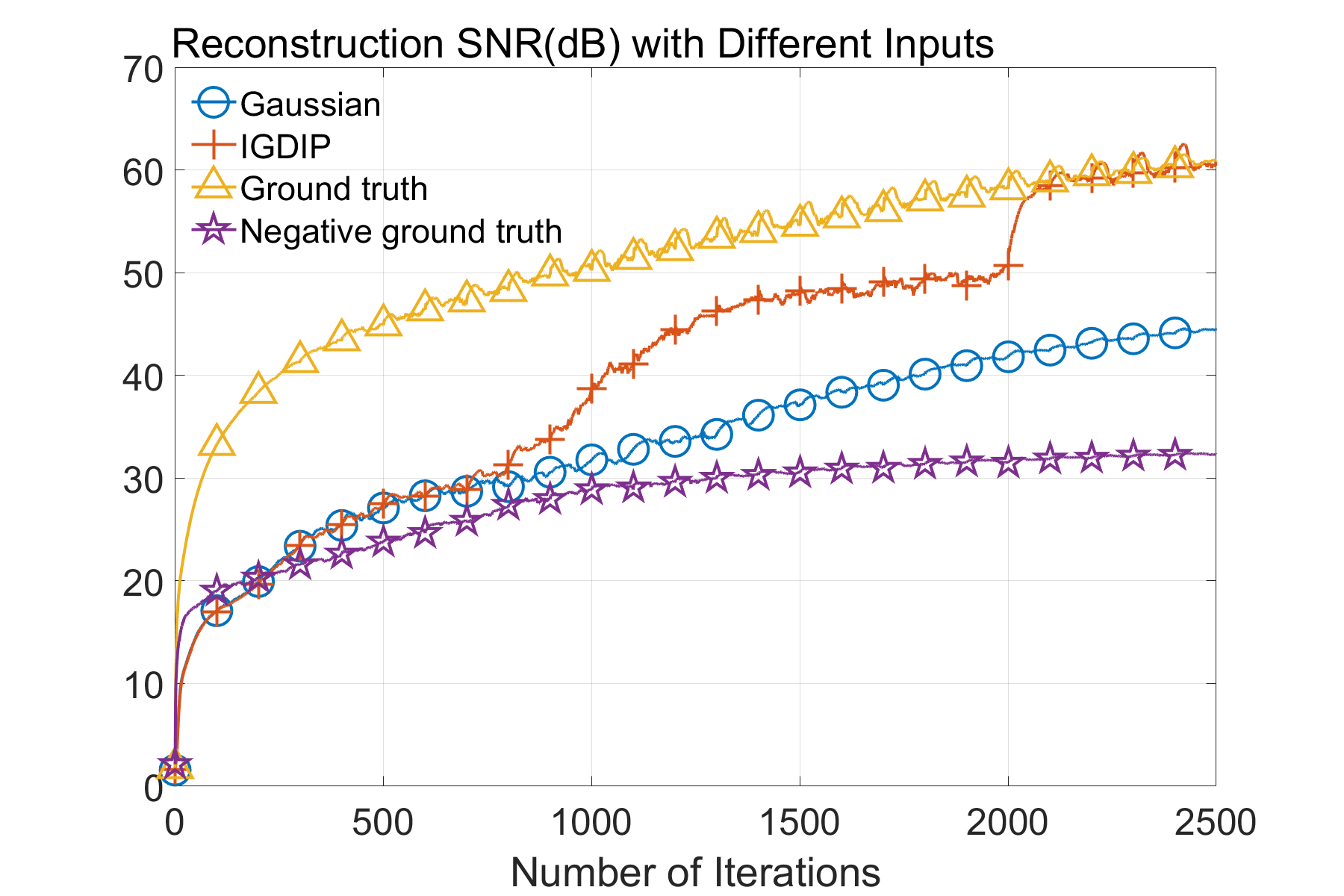}
   \caption{The reconstruction SNR under different numbers of iterations with different inputs.}
   \label{fig:different inputs}
\end{figure}

To further substantiate the impact of the input, we conducted a repeat of the above task using four distinct network inputs. These inputs include the ground truth $\boldsymbol x$, the negative of ground truth $(1-\boldsymbol x)$, Gaussian white noise $\mathcal{N}(0,1)$, and our proposed SDIP method which will be introduced later. The corresponding result is shown in Fig.\ref{fig:different inputs}, where using ground truth as network input results in the best accuracy, and using the negative ground truth as the network input leads to the worst accuracy.

We have also tried different operators $\boldsymbol{\rm H}$ for the above experiments, and obtained similar results.

\begin{figure*}
	\centering
	\begin{minipage}[b]{.19\linewidth}
		\centering
		\centerline{\includegraphics[width=\linewidth]{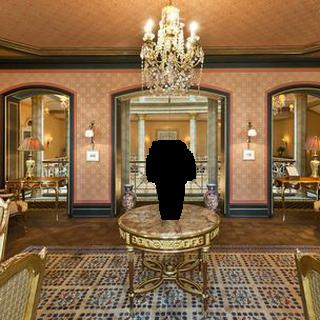}}
		\centerline{(a)}\medskip
	\end{minipage}
	\begin{minipage}[b]{.19\linewidth}
		\centering
		\centerline{\includegraphics[width=\linewidth]{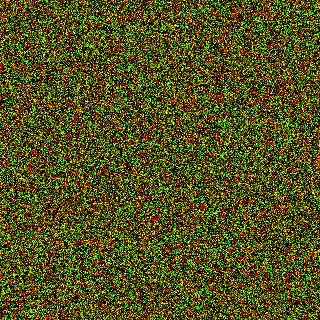}}
		\centerline{(b)}\medskip
	\end{minipage}
	\begin{minipage}[b]{.19\linewidth}
		\centering
		\centerline{\includegraphics[width=\linewidth]{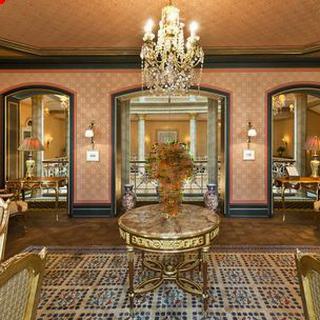}}
		\centerline{(c)}\medskip
	\end{minipage}
	\begin{minipage}[b]{.19\linewidth}
		\centering
		\centerline{\includegraphics[width=\linewidth]{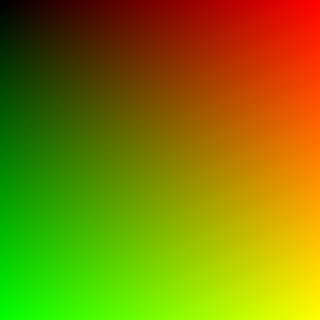}}
		\centerline{(d)}\medskip
	\end{minipage}
	\begin{minipage}[b]{.19\linewidth}
		\centering
		\centerline{\includegraphics[width=\linewidth]{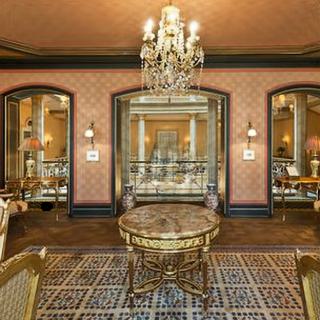}}
		\centerline{(e)}\medskip
	\end{minipage}
	\vspace{-0.4cm}
	\caption{Image inpainting using different inputs. (a) the inpainting image; (b) using uniformly distributed random vector as network input, the corresponding inpainting result is (c); (d) using an image with constant gradient as network input, the corresponding inpainting result is (e).}
	\label{inpainting}
\end{figure*}

Another example demonstrating the impact of the network input is available from the original DIP paper~\cite{ulyanov2018deep}. In the task of inpainting Fig.\ref{inpainting}a, the authors deviated from the use of a random vector (Fig.\ref{inpainting}b) and instead opted for Fig.\ref{inpainting}d as the network input. This choice may be attributed to the spatial continuity of colors between adjacent pixels in the input image, which likely aids the network in utilizing features that are closer in spatial proximity. The inpainting results (Fig.\ref{inpainting}c and Fig.\ref{inpainting}e) corresponding to the two network inputs further validate this assumption.

All these observations demonstrate that, despite the use of a randomly initialized neural network in DIP related methods, the network input can still greatly impact the final result. The more similar the input is to the ground truth, the more accurately the network can perform image recovery. However, in many cases, obtaining an input image similar to the desired result is challenging. Moreover, an inappropriate input image may even induce severe artifacts and downgrade the network output.

\begin{figure}[htb]
    \centering
    \hfill
    \begin{minipage}[b]{.45\linewidth}
        \centering
        \centerline{\includegraphics[width=\linewidth]{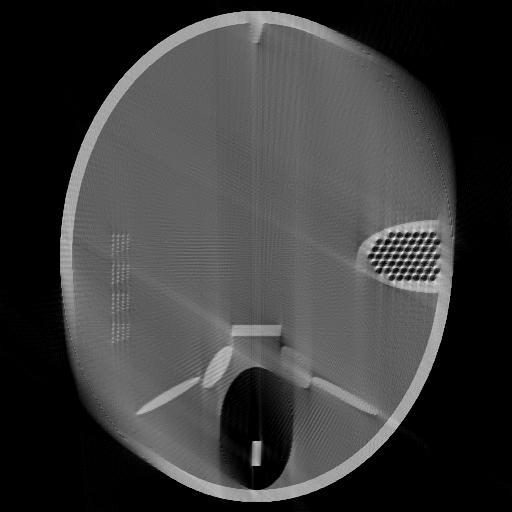}}
        \centerline{(a) Steepest Descent (23.29dB)}\medskip
    \end{minipage}
    \hfill
    \begin{minipage}[b]{.45\linewidth}
        \centering
        \centerline{\includegraphics[width=\linewidth]{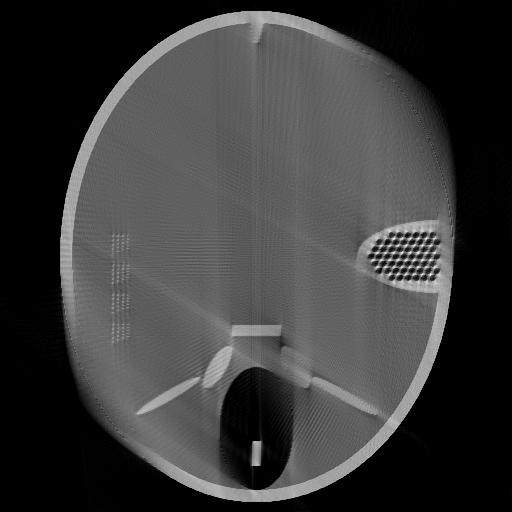}}
        \centerline{(b) CDIP (19.82dB)}\medskip
    \end{minipage}
    
    \hfill
    \begin{minipage}[b]{.45\linewidth}
        \centering
        \centerline{\includegraphics[width=\linewidth]{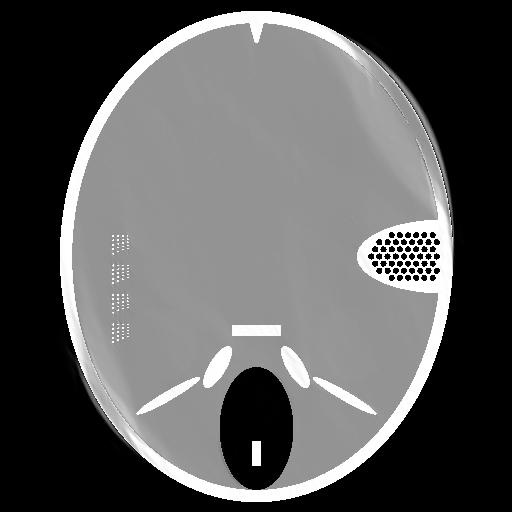}}
        \centerline{(c) DIP (29.88dB)}\medskip
    \end{minipage}
    \hfill
    \begin{minipage}[b]{.45\linewidth}
        \centering
        \centerline{\includegraphics[width=\linewidth]{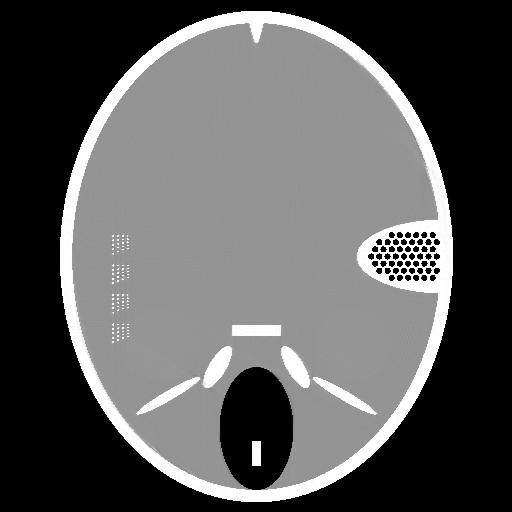}}
        \centerline{(d) SDIP (38.61dB)}\medskip
    \end{minipage}
    \caption{The limited-angle ($0^\circ$ to $120^\circ$ $1$ view per degree) CT reconstruction of Forbild phantom. (a) Steepest Descent; (b) CDIP, using (a) as network input; (c) DIP, using Gaussian random vector as network input; (d) the proposed SDIP method.}
    \label{dipcomparison}
\end{figure}

Fig.\ref{dipcomparison} shows the limited-angle CT reconstruction results of the Forbild phantom using different methods, where the projection angle goes from $0^\circ$ to $120^\circ$ with $1^\circ$ increments. Fig.\ref{dipcomparison}a presents the reconstruction result of the steepest descent algorithm. Due to the missing projections, the reconstruction result suffers from severe artifacts. Fig.\ref{dipcomparison}b shows the result of the CDIP algorithm using Fig.\ref{dipcomparison}a as input. It is evident that CDIP cannot remove the existing artifacts in Fig.\ref{dipcomparison}a and, in fact, further degrades the reconstruction accuracy. On the contrary, the output of the original DIP method (Fig.\ref{dipcomparison}c) significantly suppresses the artifacts and achieves a much better result. Finally, Fig.\ref{dipcomparison}d shows the reconstruction result of our proposed method (SDIP), which is very close to the ground truth. Its details will be introduced in Section \ref{sec:sdip}.

\subsection{Correlation of Variations between Input and Output}
\label{sec:IOrelation}
CDIP utilizes the correlation between the network input and corresponding output. However, as mentioned before, a good reference image may be unobtainable in many applications. To address this problem, we further propose the following properties: the changes in the output strongly correlate with the changes in the input in each DIP optimization iteration. 

\begin{figure*}
	\centering
	\begin{minipage}[b]{.32\linewidth}
		\centering
		\centerline{\includegraphics[width=\linewidth]{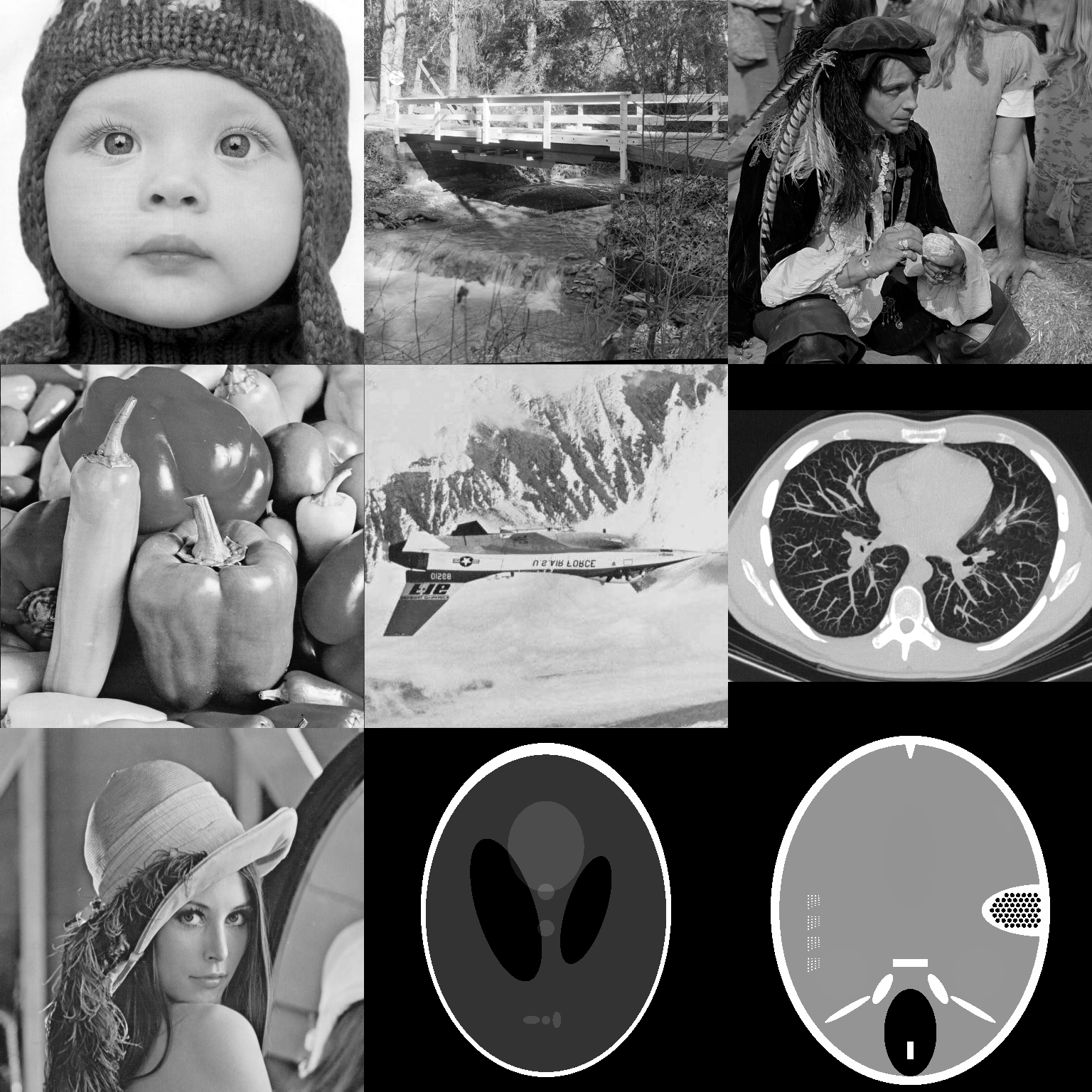}}
	\end{minipage}
	\begin{minipage}[b]{.16\linewidth}
		\begin{minipage}[b]{1\linewidth}
		\centering
		\centerline{\includegraphics[width=\linewidth]{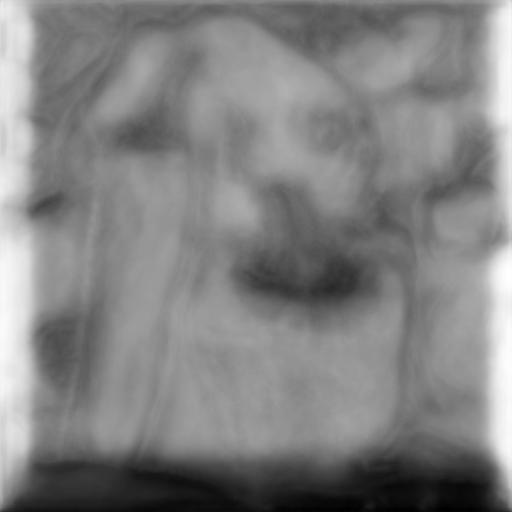}}
	\end{minipage}
 
    \begin{minipage}[b]{1\linewidth}
		\centering
		\centerline{\includegraphics[width=\linewidth]{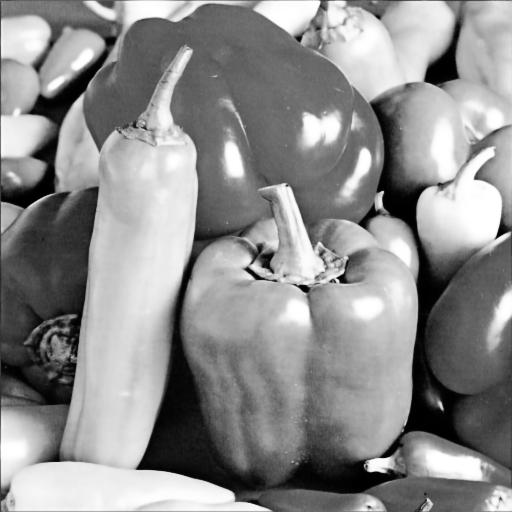}}
	\end{minipage}
	\end{minipage}
	\begin{minipage}[b]{.16\linewidth}
		\begin{minipage}[b]{1\linewidth}
		\centering
		\centerline{\includegraphics[width=\linewidth]{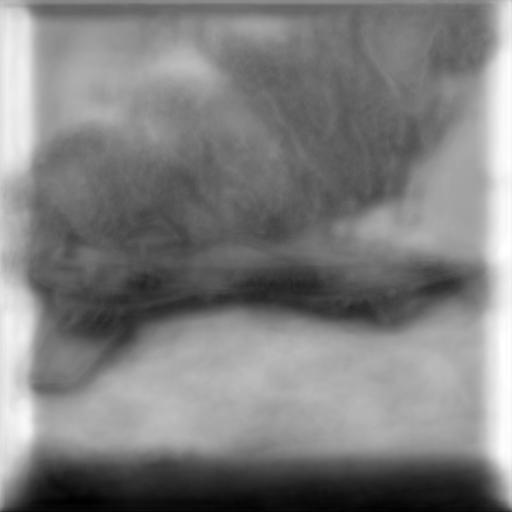}}
	\end{minipage}
 
    \begin{minipage}[b]{1\linewidth}
		\centering
		\centerline{\includegraphics[width=\linewidth]{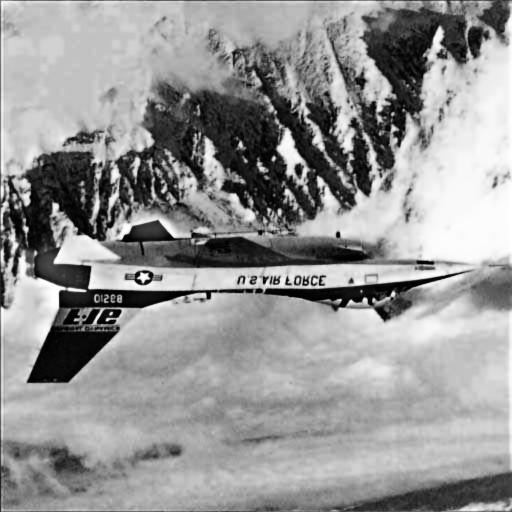}}
	\end{minipage}
	\end{minipage}
	\begin{minipage}[b]{.16\linewidth}
		\begin{minipage}[b]{1\linewidth}
		\centering
		\centerline{\includegraphics[width=\linewidth]{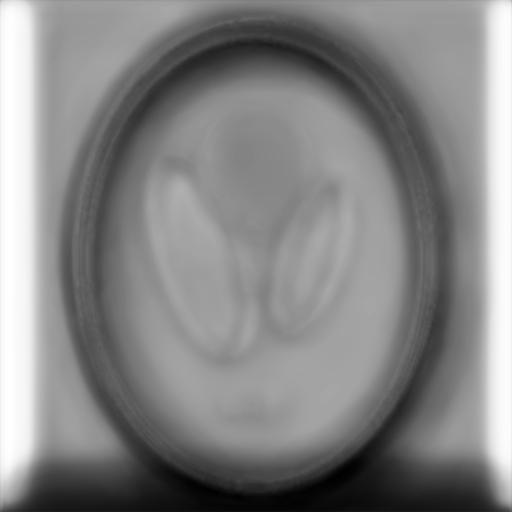}}
	\end{minipage}
 
    \begin{minipage}[b]{1\linewidth}
		\centering
		\centerline{\includegraphics[width=\linewidth]{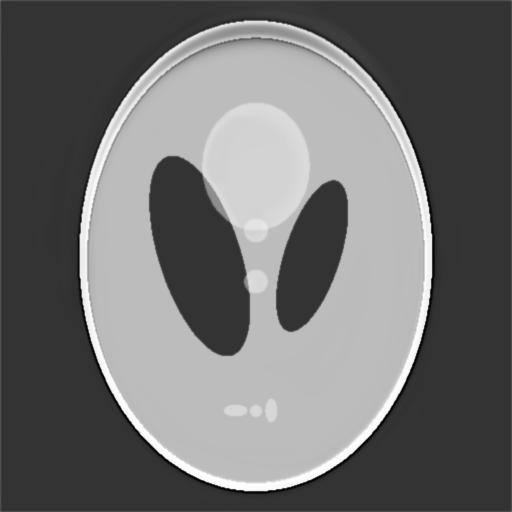}}
	\end{minipage}
	\end{minipage}
	\begin{minipage}[b]{.16\linewidth}
		\begin{minipage}[b]{1\linewidth}
		\centering
		\centerline{\includegraphics[width=\linewidth]{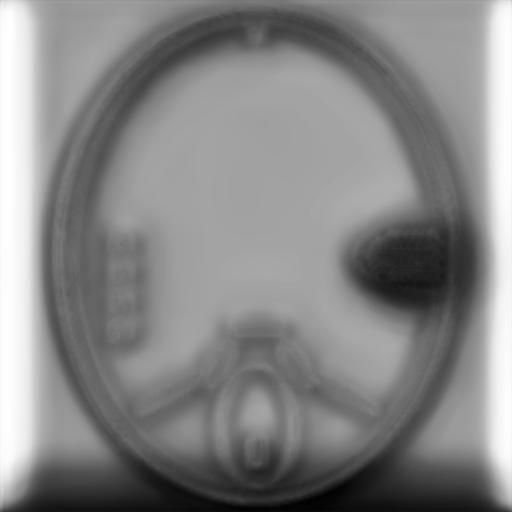}}
	\end{minipage}
 
    \begin{minipage}[b]{1\linewidth}
		\centering
		\centerline{\includegraphics[width=\linewidth]{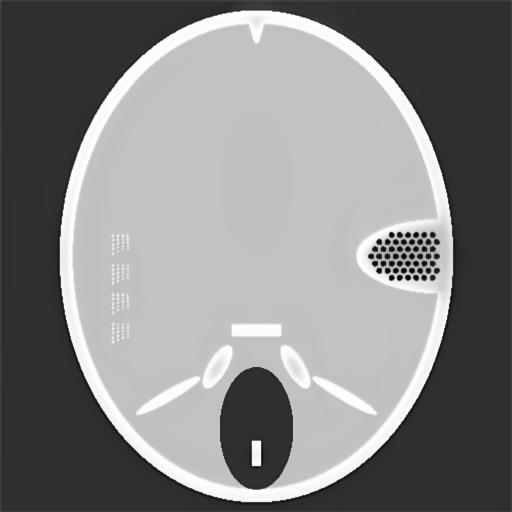}}
	\end{minipage}
	\end{minipage}
 
	\caption{The strong correlation between the change of network input and the corresponding output. The $9$ images on the left are used in the experiments in Section \ref{sec:IOrelation}. The images on the right are the mean output changes when the network input is replaced by the fourth, fifth, eighth, and ninth images on the left. The first row corresponds to the DIP method, and the second row corresponds to the CDIP method. }
	\label{IOrelation}
\end{figure*}

To verify this idea, the following experiments are conducted: we use Equation \ref{cdipexp} to recover one of the nine images shown on the left of Fig.\ref{IOrelation}. For the DIP method, the network input $\boldsymbol z$ is set to a Gaussian random vector, and for the CDIP method, $\boldsymbol z$ is set to the target image directly. Then, at the $1$st, $300$th, $600$th, ..., $1500$th iteration, the input $\boldsymbol z$ is replaced by one of the remaining eight images, and the cosine similarity between the changes of the network input and output is calculated. Such experiments are repeated $1000$ times, so that the mean cosine similarity and mean output changes can be obtained.

Table \ref{cosdistance} shows the average cosine similarities between the changes of input and output at different iterations for the DIP and the CDIP algorithms. Both the DIP and CDIP achieve very high cosine similarity at any iteration. For ease of understanding, some of the mean output changes are presented on the right part of Fig.\ref{IOrelation}, where the first row corresponds to the DIP algorithm, and the second corresponds to the CDIP algorithm. It is apparent that these images, especially those corresponding to the CDIP method, are similar to their respective counterpart.

Based on these observations, we reach the following conclusion: in DIP networks, the changes in the input can statistically nearly identically affect the corresponding output. In other words, the following equation holds true statistically in any DIP iteration:
\begin{equation}
|G(\boldsymbol \theta |\boldsymbol z_1) - G(\boldsymbol \theta |\boldsymbol z_2)| \approx \alpha |\boldsymbol z_1 - \boldsymbol z_2|.
\end{equation}

\begin{table}[htbp]
  \centering
  \caption{The similarity between the modified input and corresponding output of DIP and CDIP methods.}
  \setlength{\tabcolsep}{0.5mm}{}
  \begin{tabular}{@{}lcccccc@{}}
  \toprule
  \multicolumn{7}{c}{Cosine Distance} \\
    \toprule
    Number of Iterations & 1 & 300 & 600 & 900 & 1200 & 1500\\
    \midrule
    DIP & 0.79 & 0.71 & 0.72 & 0.71 & 0.72 & 0.72\\
    CDIP & 0.93 & 0.94 & 0.94 & 0.94 & 0.94 & 0.94\\
    \bottomrule
  \end{tabular}
  \label{cosdistance}
\end{table}

\subsection{Self-Reinforcement DIP}
\label{sec:sdip}
\begin{figure}[htbp]
  \centering
   \includegraphics[width=1\linewidth]{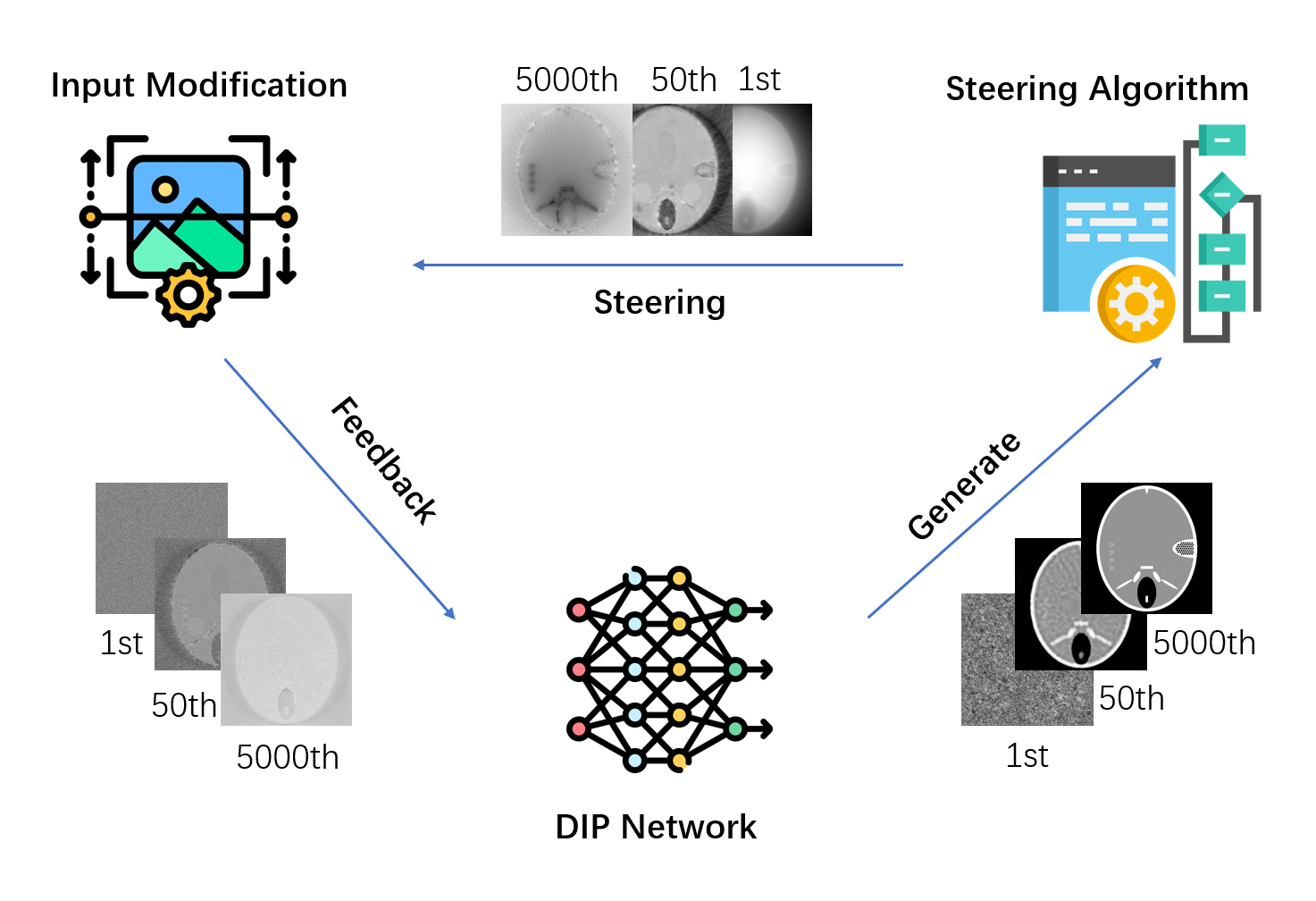}
\vspace{-0.8cm}
   \caption{The flowchart of the SDIP framework. The images on the left indicate the change in the network input during the iteration, the images on the right indicate the network output, and the images on the top illustrate the modifications that will be applied to the network input.} 
   \label{framework}
\end{figure}

In the above section, we have shown that the changes in the input can statistically nearly identically affect the corresponding output. Based on this, we propose the Self-Reinforcement DIP (SDIP) framework. Its basic idea is to steer the algorithm to a more reasonable result by iteratively adjusting its input. A brief flowchart is shown in Fig.\ref{framework}, which contains the following three main steps.

\vspace{-0.8cm}

\subsubsection{DIP Network for Image Generation}
SDIP still utilizes the DIP network for image generation. In this paper, we use the standard U-net architecture as a prior, primarily due to its widespread use across various image processing problems. We believe it possesses sufficient versatility to handle multiple image recovery tasks. However, this does not imply that SDIP is only limited to U-net. We have also experimented with other network structures, especially those mentioned in the original DIP paper~\cite{ulyanov2018deep}, and have achieved similar improvements.

\subsubsection{Steering Algorithm for Introducing Additional Prior}
The key difference between SDIP and the original DIP is the use of the steering algorithm. Its objective is to guide the DIP network by modifying the network input based on the previous iteration's network output, leveraging the properties mentioned in Section \ref{sec:IOrelation}. The steering algorithm can also be regarded as an additional prior, collaborating with DIP to achieve better results. In that case, it can be any algorithm that can provide meaningful information to the network input.

Assuming the ground truth image ($x_{\rm gt}$) is known, the steering algorithm can directly compute the residual between the current network output ($x$) and $x_{\rm gt}$, which represents the ideal scenario for the optimal steering algorithm. In practice, although ground truth images or reference images are unobtainable, we can employ existing algorithms, such as linear optimization methods or pre-trained neural network methods, as steering algorithms to guide the DIP network. These algorithms, when used independently, often fail to produce high-quality results due to the highly ill-posed nature of the given inverse problems (Fig.\ref{dipcomparison}a). However, when serving as steering algorithms, the DIP network can generate high-quality preliminary results for these methods (Fig.\ref{dipcomparison}c), thereby enhancing their effectiveness and finally enhancing the overall performance of the SDIP framework (Fig.\ref{dipcomparison}d).

In this paper, for simplicity, only the simple linear gradient descent algorithm minimizing the data inconsistency is used as the steering algorithm. In that case, such an SDIP method can take advantage of both DIP and the prior integrated in the gradient descent method. For example, when the inverse problem is ill-posed, SDIP can first utilize the DIP to generate a reasonable initial guess, and then utilize the steering algorithm to refine it, resulting in a better outcome.

\subsubsection{Input Modification with Dynamic Step Size}
Although SDIP tries to obtain improved results through the iterative modification of the network input, it should be noted that significant variations in the input to the DIP network may compromise its stability and performance. Furthermore, in the early iterations when DIP has not yet generated sufficiently good preliminary results, steering algorithms may fail to effectively guide or even mislead the DIP network. Therefore, it is necessary to regulate the magnitude of updates to the network input. We aim to suppress the steering algorithm during the early iterations, allowing it to take effect after the DIP network has generated satisfactory preliminary results. In this paper, the network input $\boldsymbol z$ undergoes a slight adjustment by a vector of magnitude $\frac{10^{-3}}{1+e^{-(\frac{n-n_c}{n_s})}}$. It is a modified sigmoid function centered at $n_c$ and stretched by a factor of $n_s$. In that case, the modification from the steering algorithm will be suppressed before the $n_c$-th iteration and subsequently augmented in later iterations. Finally, at the beginning of each iteration, we normalize the modified network input, ensuring that the network's input $\boldsymbol z$ remains a unit vector, with only its direction changing throughout each iteration.

\subsubsection{SDIP Workflow}
To summarize, the details of the SDIP are shown in Algorithm \ref{alg1}. In each iteration, the network input will be normalized to a unit vector (line 3). A randomly initialized convolutional neural network is employed to generate a temporal result $\boldsymbol x$ (line 4). An objective function such as data inconsistency can be utilized to update the network $G$ (line 5). Meanwhile, the steering algorithm is employed to modify the next iteration's network input based on the current network output. Here, the basic gradient descent algorithm is employed as the steering algorithm. Consequently, the objective function's derivative with respect to $\boldsymbol x$, denoted as $\boldsymbol r = \boldsymbol{\rm H}^{\rm T} ( \boldsymbol y - \boldsymbol{\rm H} \boldsymbol x)$, will be utilized to adjust the network input (line 6 and line 7).

\begin{algorithm}[t]
	\caption{Self-Reinforcement Deep Image Prior}
	\label{alg1}
	\hspace*{0.02in} {\bf Input:} 
	measure matrix $\boldsymbol{\rm H}$, measurement $\boldsymbol y$, as well as the two hyper parameters $n_c$ and $n_s$. $G(\boldsymbol \theta| \boldsymbol z)$ indicates an untrained convolutional neural network as well as its corresponding weight $\boldsymbol \theta$ and input  $\boldsymbol z$.
	\begin{algorithmic}[1]
		\State randomly initiate $\boldsymbol z$
		\State {\bf Repeat:} 
        \State \quad $\boldsymbol z = \frac{\boldsymbol z}{||\boldsymbol z||_2^2} $
        \State \quad $\boldsymbol x = G(\boldsymbol \theta| \boldsymbol z)$
        \State \quad using the loss function $||\boldsymbol y - \boldsymbol{\rm H}\boldsymbol x||_2^2$ to update the network $G$'s weights $\boldsymbol \theta$
		\State \quad $\boldsymbol r$ = $\boldsymbol{\rm H}^{\rm T} ( \boldsymbol y - \boldsymbol{\rm H} \boldsymbol x)$
		\State \quad $\boldsymbol z = \boldsymbol z + \frac{10^{-3}}{1+e^{-(\frac{n-n_c}{n_s})}} \frac{\boldsymbol r}{||\boldsymbol r||_2^2}$
		\State {\bf Until:} convergence, or fixed number of iterations is reached.  
	\end{algorithmic}
\end{algorithm}

\section{Applications}
\label{sec:app}
In this section, the proposed method will be compared with the original DIP method and other conventional methods across various applications, including CT reconstruction, deblurring, and super-resolution. The objective is to show that the SDIP method is superior to the DIP method in most cases.

\subsection{Selection of the Hyperparameters}
The proposed method utilizes two hyperparameters, $n_c$ and $n_s$, to control the strength of updates across various iterations. A series of experiments were conducted to identify an optimal combination of both hyperparameters: given the total number of iterations is $10000$, we settled on a range of $0$ to $10000$ and a range of $100$ to $1000$ for $n_c$ and $n_s$ respectively. The first experiment corresponds to a limited-angle CT reconstruction using the angular range between $0^\circ$ and $135^\circ$. In that case, the corresponding inverse problem is highly ill-posed. The second and third experiments correspond to CT reconstruction using the angular ranges between $0^\circ$ and $180^\circ$ as well as $0^\circ$ and $360^\circ$, the corresponding inverse problem is well-posed and over-determined respectively. The corresponding reconstruction SNR is shown in Table.\ref{tab:135}, Table.\ref{tab:180}, and Table.\ref{tab:360}. 

The results shown in the aforementioned tables indicate that the selection of $n_c$ mainly depends on the nature of the inverse problem itself. In instances where the inverse problem is highly ill-posed (Table.\ref{tab:135}), conventional iterative reconstruction (IR) methods implemented in the steering algorithm may result in pronounced artifacts. In such cases, it is advisable to use a larger $n_c$. This adjustment gives the algorithm more time to leverage the properties of the deep image prior (DIP) for obtaining a better initial guess. Conversely, if the inverse problem is closer to being well-posed (Table.\ref{tab:180}) or over-determined (Table.\ref{tab:360}), a smaller $n_c$ enables the steering algorithm to take in the optimization process more promptly. As for $n_c$, it is imperative for $n_c$ to be sufficiently large to yield a smooth function, preventing significant disruptions to the network stability caused by input updates. 

In our experiments, for simplicity, we have set the parameters to $n_s = 500$, $n_c = 5000$, and the total number of iterations is $10000$. By employing these values, the steering algorithm is suppressed during the first half of the reconstruction process and gradually augmented during the second half.

\begin{table}[H]
\centering
\caption{CT Reconstruction SNR(dB) with the projection angular range $0^\circ-135^\circ$, the system is highly ill-posed.}
\begin{tabular}{|l|c|c|c|c|}
\hline
\diagbox{$n_c$}{$n_s$}      & 100 & 200 & 500 & 1000\\
\hline
0    & 26.72 & 27.64 & 26.95 & 28.09  \\
\hline
2500 & 31.98 & 31.85 & 32.78 & 31.96  \\
\hline
5000 & 28.68 & 33.99 & 34.02 & 34.16  \\
\hline
7500 & 34.83 & 34.73 & 40.23 & 38.94  \\
\hline
10000 & 38.41 & 41.29 & 39.95 & 39.83  \\
\hline
\end{tabular}
\label{tab:135}
\end{table}

\begin{table}
\centering
\caption{CT Reconstruction SNR(dB) with the projection angular range $0^\circ-180^\circ$, the system is well-posed.}
\begin{tabular}{|l|c|c|c|c|}
\hline
\diagbox{$n_c$}{$n_s$}      & 100 & 200 & 500 & 1000\\
\hline
0    & 45.36 & 45.13 & 45.54 & 45.12  \\
\hline
2500 & 45.96 & 45.39 & 46.51 & 46.34  \\
\hline
5000 & 46.93 & 46.35 & 46.65 & 47.15  \\
\hline
7500 & 45.51 & 45.77 & 46.80 & 47.43  \\
\hline
10000 & 44.02 & 45.52 & 45.62 & 46.52  \\
\hline
\end{tabular}
\label{tab:180}
\end{table}

\begin{table}
\centering
\caption{CT Reconstruction SNR(dB) with the projection angular range $0^\circ-360^\circ$, the system is over-determined.}
\begin{tabular}{|l|c|c|c|c|}
\hline
\diagbox{$n_c$}{$n_s$}      & 100 & 200 & 500 & 1000\\
\hline
0    & 53.93 & 53.20 & 54.41 & 53.84  \\
\hline    
2500 & 54.44 & 54.33 & 53.37 & 55.15  \\
\hline
5000 & 53.74 & 51.79 & 54.36 & 54.94  \\
\hline
7500 & 51.35 & 51.63 & 52.37 & 52.82  \\
\hline
10000 & 47.80 & 47.19 & 47.57 & 48.94  \\
\hline
\end{tabular}
\label{tab:360}
\end{table}

\begin{figure*}
	\begin{minipage}[b]{.16\linewidth}
		\centering
		\centerline{\includegraphics[width=\linewidth]{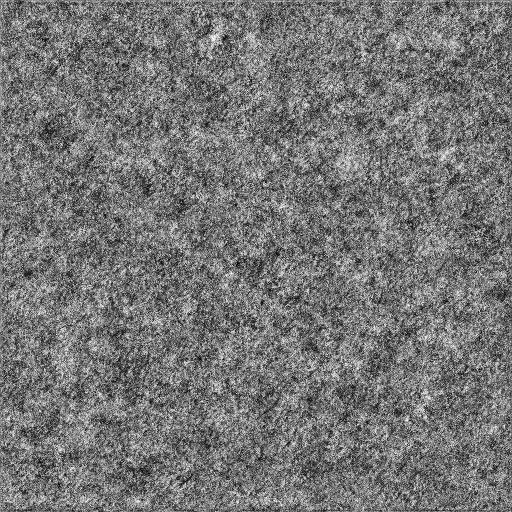}}
	\end{minipage}
	\begin{minipage}[b]{.16\linewidth}
		\centering
		\centerline{\includegraphics[width=\linewidth]{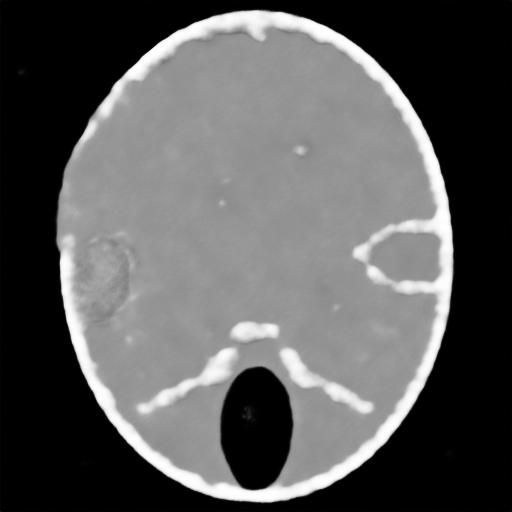}}
	\end{minipage}
	\begin{minipage}[b]{.16\linewidth}
		\centering
		\centerline{\includegraphics[width=\linewidth]{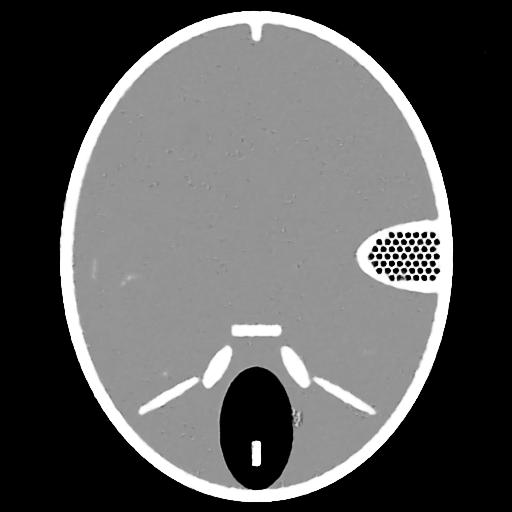}}
	\end{minipage}
	\begin{minipage}[b]{.16\linewidth}
		\centering
		\centerline{\includegraphics[width=\linewidth]{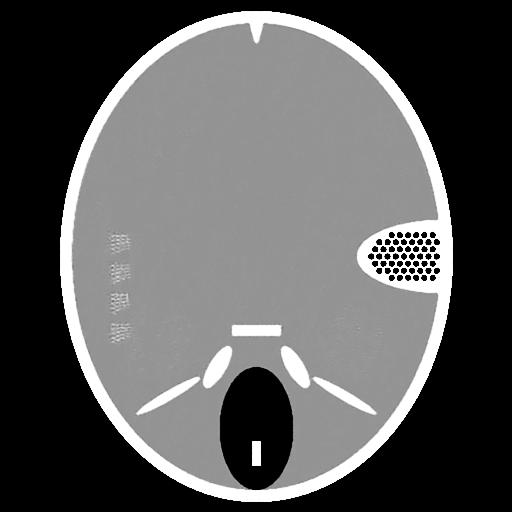}}
	\end{minipage}
	\begin{minipage}[b]{.16\linewidth}
		\centering
		\centerline{\includegraphics[width=\linewidth]{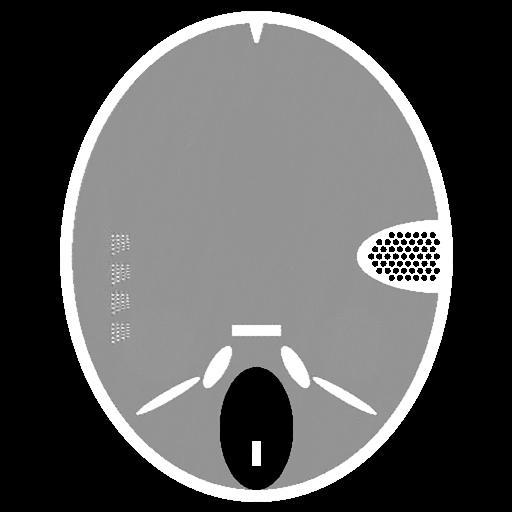}}
	\end{minipage}
	\begin{minipage}[b]{.16\linewidth}
		\centering
		\centerline{\includegraphics[width=\linewidth]{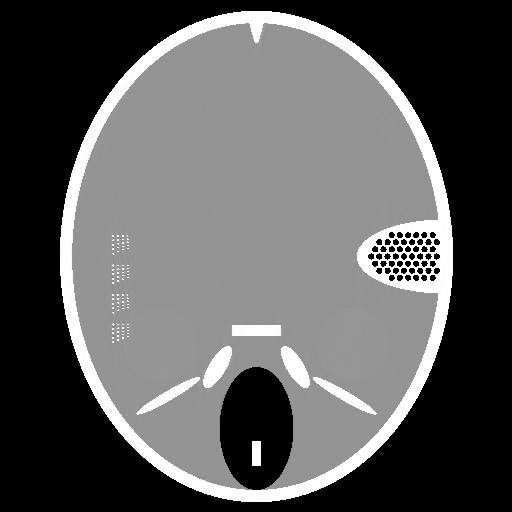}}
	\end{minipage}
 
	\begin{minipage}[b]{.16\linewidth}
		\centering
		\centerline{\includegraphics[width=\linewidth]{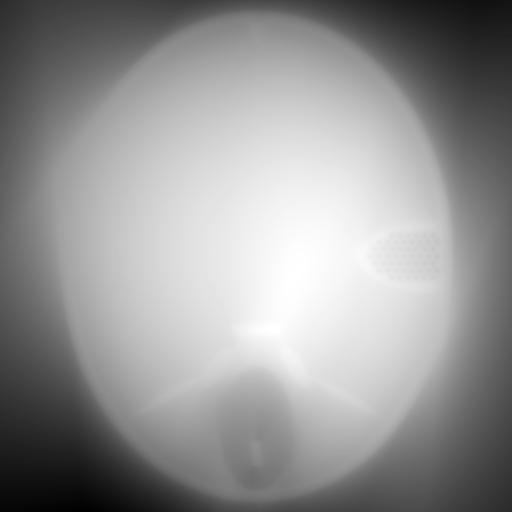}}
		\centerline{0th}\medskip
	\end{minipage}
	\begin{minipage}[b]{.16\linewidth}
		\centering
		\centerline{\includegraphics[width=\linewidth]{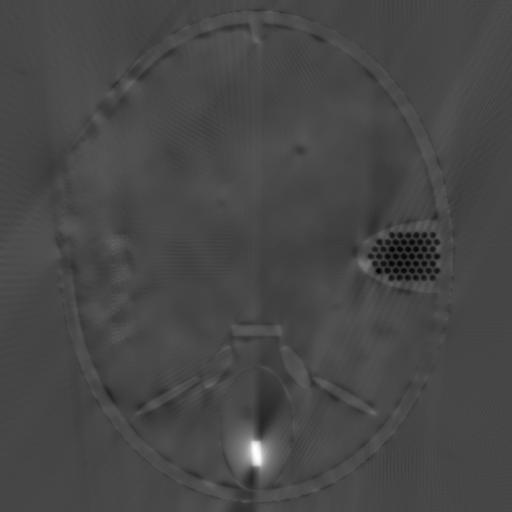}}
		\centerline{1000th}\medskip
	\end{minipage}
	\begin{minipage}[b]{.16\linewidth}
		\centering
		\centerline{\includegraphics[width=\linewidth]{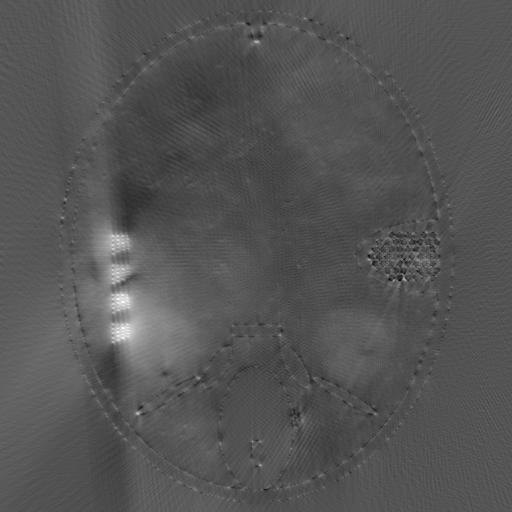}}
		\centerline{2000th}\medskip
	\end{minipage}
	\begin{minipage}[b]{.16\linewidth}
		\centering
		\centerline{\includegraphics[width=\linewidth]{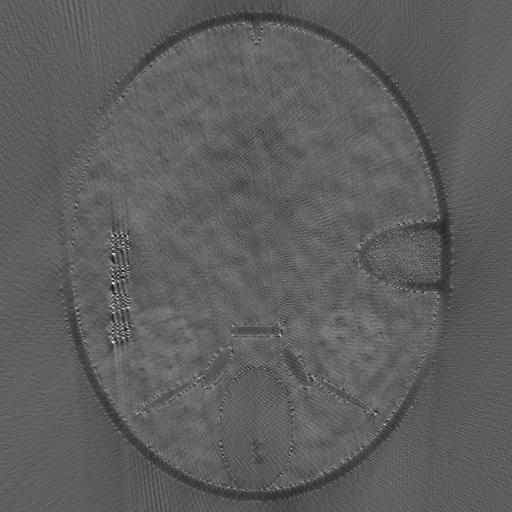}}
		\centerline{3000th}\medskip
	\end{minipage}
	\begin{minipage}[b]{.16\linewidth}
		\centering
		\centerline{\includegraphics[width=\linewidth]{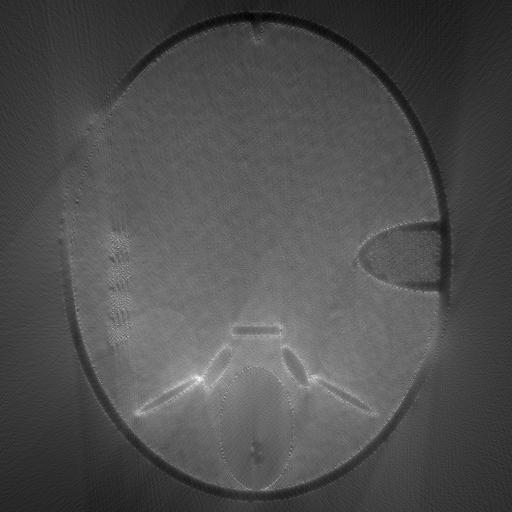}}
		\centerline{4000th}\medskip
	\end{minipage}
	\begin{minipage}[b]{.16\linewidth}
		\centering
		\centerline{\includegraphics[width=\linewidth]{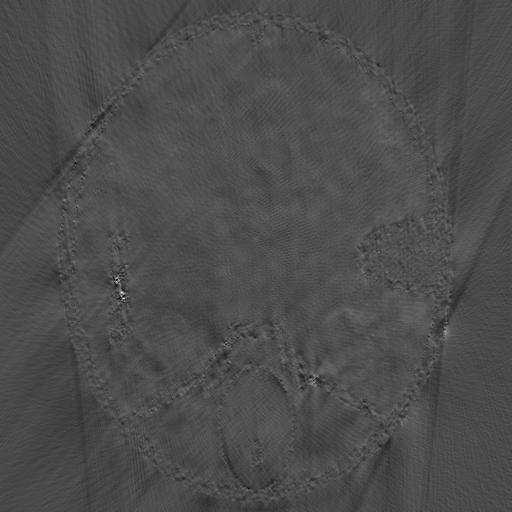}}
		\centerline{10000th}\medskip
	\end{minipage}
	\vspace{-0.4cm}
	\caption{Limited-angle CT reconstruction process of the SDIP method. The first and second rows show the output of the neural network and the steering algorithm at different iterations respectively.}
	\label{reconresultfewview}
\end{figure*}

\subsection{Computed Tomography (CT) Reconstruction}
In this section, we evaluate the SDIP framework's performance in CT reconstructions. Our goal is to reconstruct CT images under few-view and limited-angle conditions, where the corresponding inverse problem is highly under-determined. Previously, researchers proposed utilizing conventional iterative reconstruction (IR) algorithms with regularization techniques~\cite{jin2010anisotropic, shu2022global, wang2021limitedangle} or leveraging pre-trained neural network~\cite{anirudh2018lose,hu2022dior} to address this challenge. However, the effectiveness of the former approach diminishes significantly when the system is highly under-determined. The latter, while promising, hinges on the availability of extensive high-quality training datasets, which may be impractical within the domain of medical imaging~\cite{antun2020instabilities, bhadra2021hallucinations}. Recently, methods related to DIP have demonstrated significant potential in this field~\cite{shu2022sparse,baguer2020computed}. We aim to show that the proposed SDIP framework can further improve its performance.

First, to show the reconstruction process of the SDIP and the efficacy of the self-reinforcement mechanism within the SDIP framework, we conducted a limited-angle CT reconstruction experiment using SDIP. This experiment involved reconstructing images from CT scans captured over an angular range of $0^\circ - 120^\circ$, with one view per degree. The outputs of the network at various iterations, along with the outputs of the steering algorithm, are illustrated in the first and second rows of Fig.\ref{reconresultfewview}, respectively. As shown in Fig.\ref{reconresultfewview}, at the beginning (the 0th iteration) the network output is completely random as the SDIP framework utilizes a randomly initialized neural network. Then, the SDIP network's output quickly converges to the correct shape by the 1000th iteration, yet it lacks the bright rectangle at the image's lower section. This omission is detected by the steering algorithm, which then highlights this in its output. The output of the steering algorithm affects the network's input through the proposed self-reinforcement mechanism, thereby altering the network's output. Consequently, the missing rectangle is successfully reconstructed before the 2000th iteration. Similarly, at the 2000th iteration, details on the left side of the image are also missing, a flaw that the steering algorithm identifies and corrects before the 3000th iteration through the self-reinforcement mechanism. In the later iteration, the self-reinforcement mechanism continuously refines the details of the reconstructed output, yielding a high-quality outcome.

\begin{figure}[htbp]
  \centering
\includegraphics[width=1\linewidth]{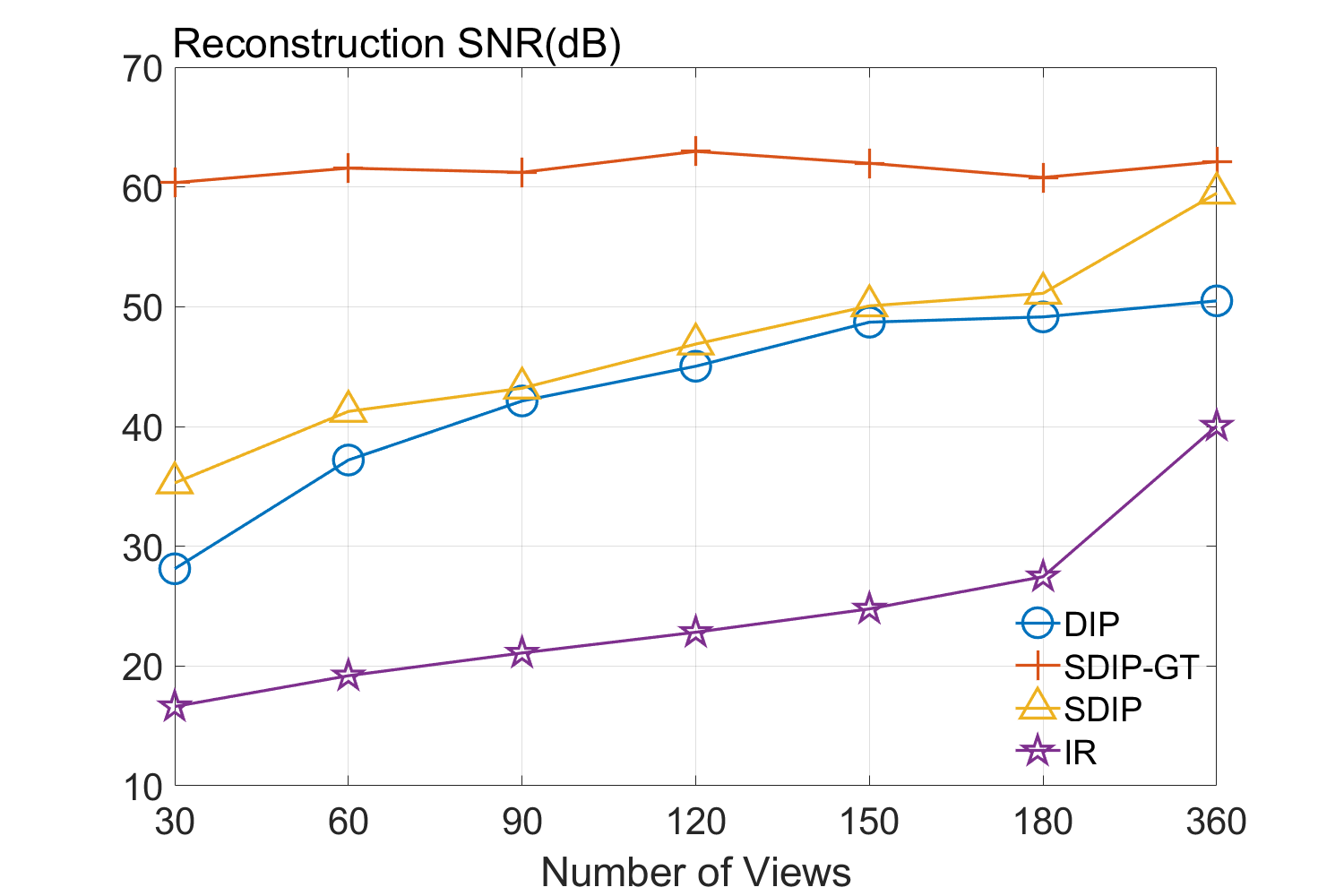}
   \caption{The reconstruction SNR of few-view CT reconstruction for different methods under different numbers of views.}
   \label{fig:snrfewview}
\end{figure}

\begin{figure}[htbp]
  \centering
\includegraphics[width=1\linewidth]{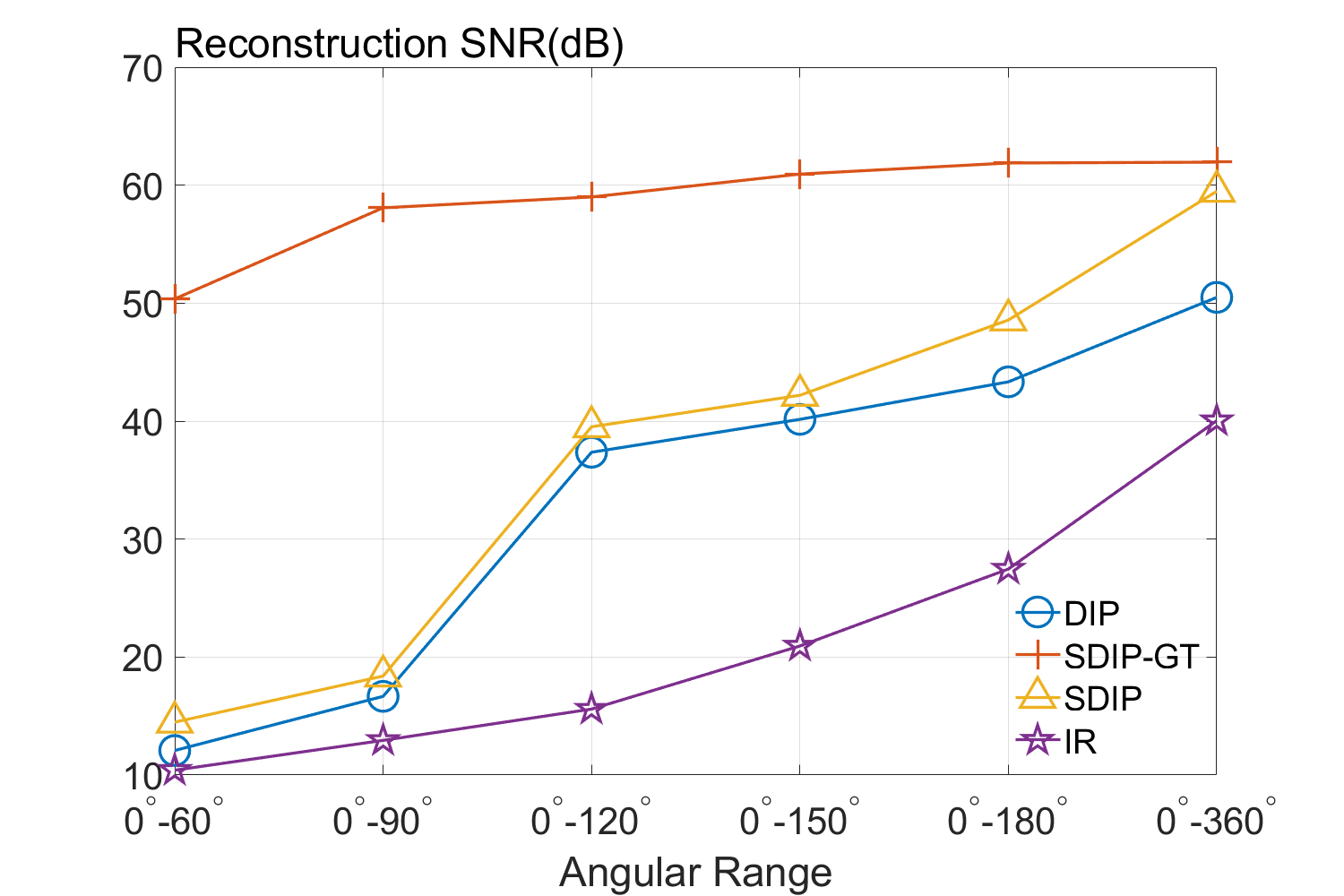}
   \caption{The reconstruction SNR of limited-angle CT reconstruction for different methods under different angular ranges.}
   \label{fig:snrlimitedangle}
\end{figure}

\begin{figure}[htbp]
	\centering
	\begin{minipage}[b]{.3\linewidth}
		\centering
		\centerline{\includegraphics[width=\linewidth]{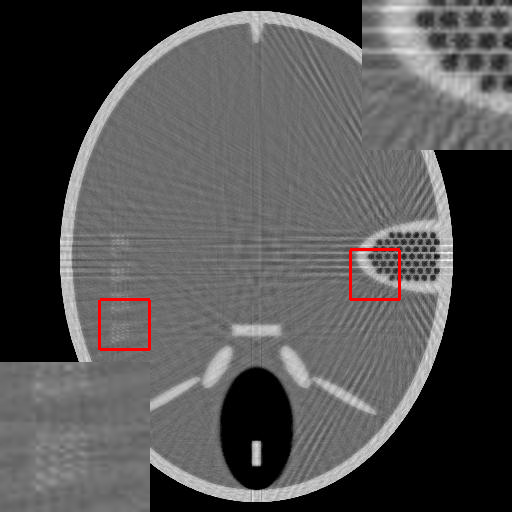}}
		\centerline{(a) 19.13dB}\medskip
	\end{minipage}
	\begin{minipage}[b]{.3\linewidth}
		\centering
		\centerline{\includegraphics[width=\linewidth]{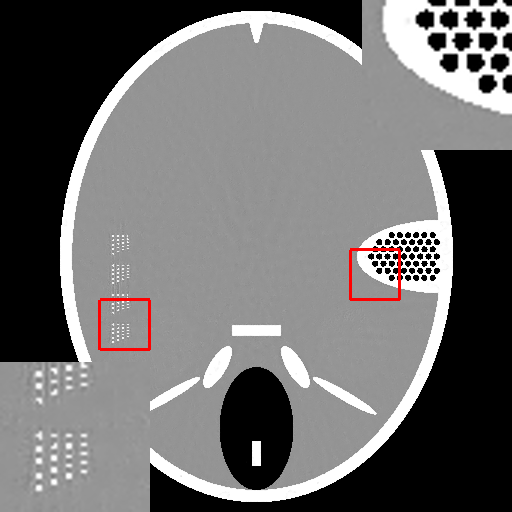}}
		\centerline{(b) 38.23dB}\medskip
	\end{minipage}
	\begin{minipage}[b]{.3\linewidth}
		\centering
		\centerline{\includegraphics[width=\linewidth]{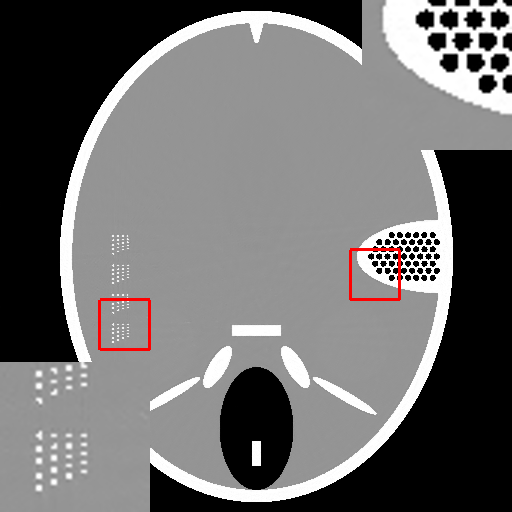}}
		\centerline{(c) 42.92dB}\medskip
	\end{minipage}

	\begin{minipage}[b]{.3\linewidth}
		\centering
		\centerline{\includegraphics[width=\linewidth]{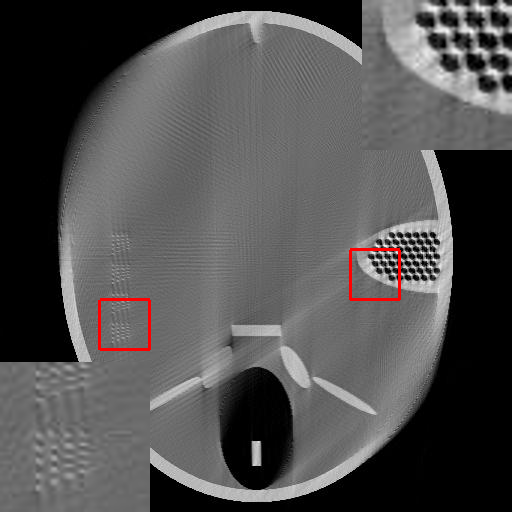}}
		\centerline{(d) 16.33}\medskip
	\end{minipage}
	\begin{minipage}[b]{.3\linewidth}
		\centering
		\centerline{\includegraphics[width=\linewidth]{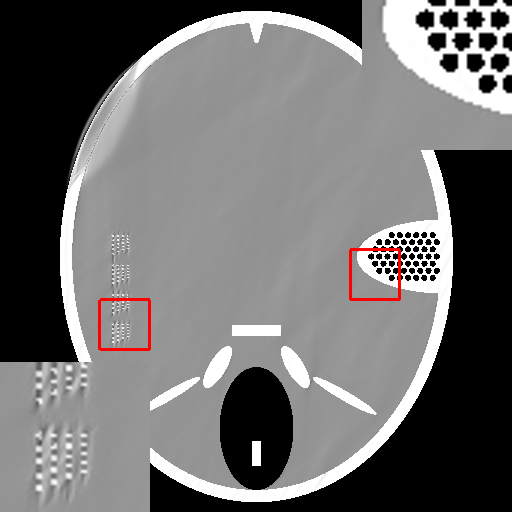}}
		\centerline{(e) 37.69dB}\medskip
	\end{minipage}
	\begin{minipage}[b]{.3\linewidth}
		\centering
		\centerline{\includegraphics[width=\linewidth]{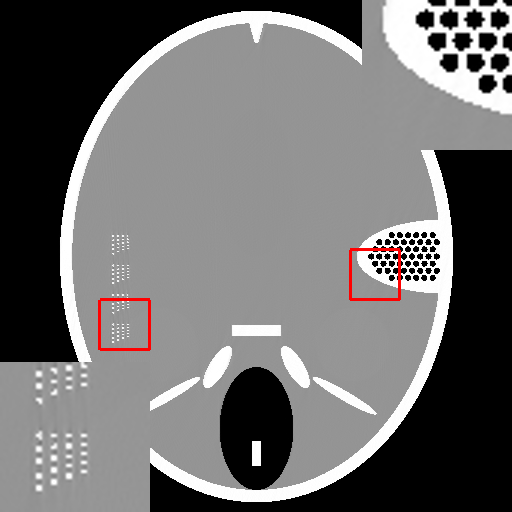}}
		\centerline{(f) 43.87dB}\medskip
	\end{minipage}

	\begin{minipage}[b]{.3\linewidth}
		\centering
		\centerline{\includegraphics[width=\linewidth]{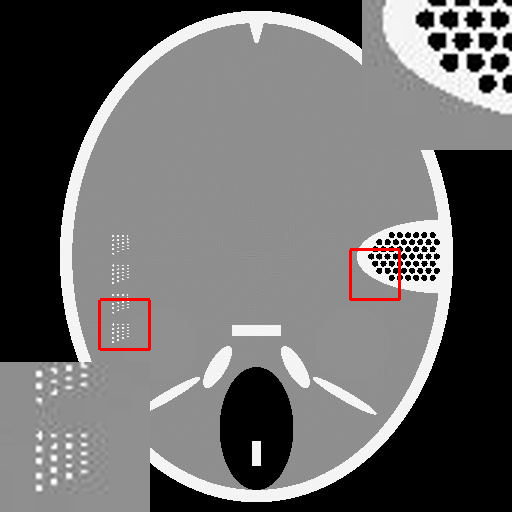}}
		\centerline{(g) 40.75dB}\medskip
		\centerline{IR}\medskip
	\end{minipage}
	\begin{minipage}[b]{.3\linewidth}
		\centering
		\centerline{\includegraphics[width=\linewidth]{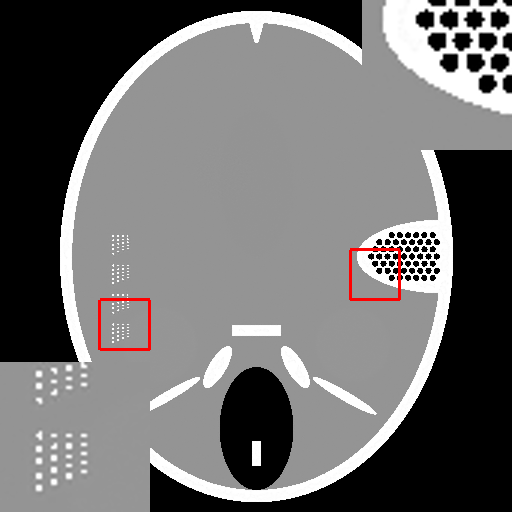}}
		\centerline{(h) 50.28dB}\medskip
		\centerline{DIP}\medskip
	\end{minipage}
	\begin{minipage}[b]{.3\linewidth}
		\centering
		\centerline{\includegraphics[width=\linewidth]{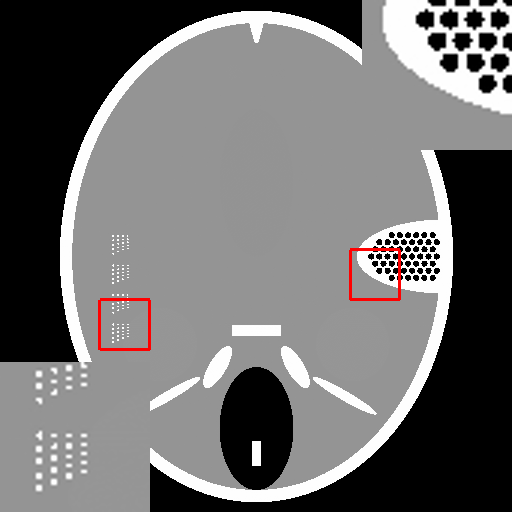}}
		\centerline{(I) 59.18dB}\medskip
  \centerline{SDIP}\medskip
	\end{minipage}
	%
	\caption{Reconstruction results of different methods under few-view (first row), limited-angle (second row), and complete-view (third row) conditions, where the first, second, and third column corresponds to conventional IR method, the original DIP method, and the proposed SDIP method respectively.}
	\label{reconresult}
\end{figure}

Then, we evaluate the performance of the SDIP framework under conditions of both few-view and limited-angle, in comparison to the original DIP method and the conventional IR method. To further verify the effectiveness of the SDIP, an additional set of experiments was conducted (termed SDIP-GT). In the SDIP-GT, it is presumed that the steering algorithm is privy to the ground truth image. Consequently, line 6 of Algorithm \ref{alg1} is modified to $\boldsymbol r = \boldsymbol x_{\rm gt} - \boldsymbol x$, where $\boldsymbol x_{\rm gt}$ represents the ground truth image. If the proposed self-reinforcement mechanism can enhance the algorithm's performance effectively, then SDIP-GT, which utilizes ground truth images to steer the algorithm, should achieve superior accuracy. The experiment results are shown in Fig.\ref{fig:snrfewview} and Fig.\ref{fig:snrlimitedangle}, where SDIP-GT maintains the highest performance and is nearly immune to the decreased number of measurements. SDIP outperforms the original DIP and the IR method. The reconstruction results of Forbild phantom~\cite{yu2012simulation} under few-view, limited-angle, and complete-view conditions are illustrated in Fig.\ref{reconresult}, it is evident that the conventional IR method suffers from severe artifacts. The original DIP approach substantially mitigates these artifacts, achieving notable improvements. The proposed SDIP framework achieves the best reconstruction results which are nearly artifact-free.

\subsection{Deblurring}
\begin{figure*}
	\centering
	\begin{minipage}[b]{.19\linewidth}
		\centering
		\centerline{\includegraphics[width=\linewidth]{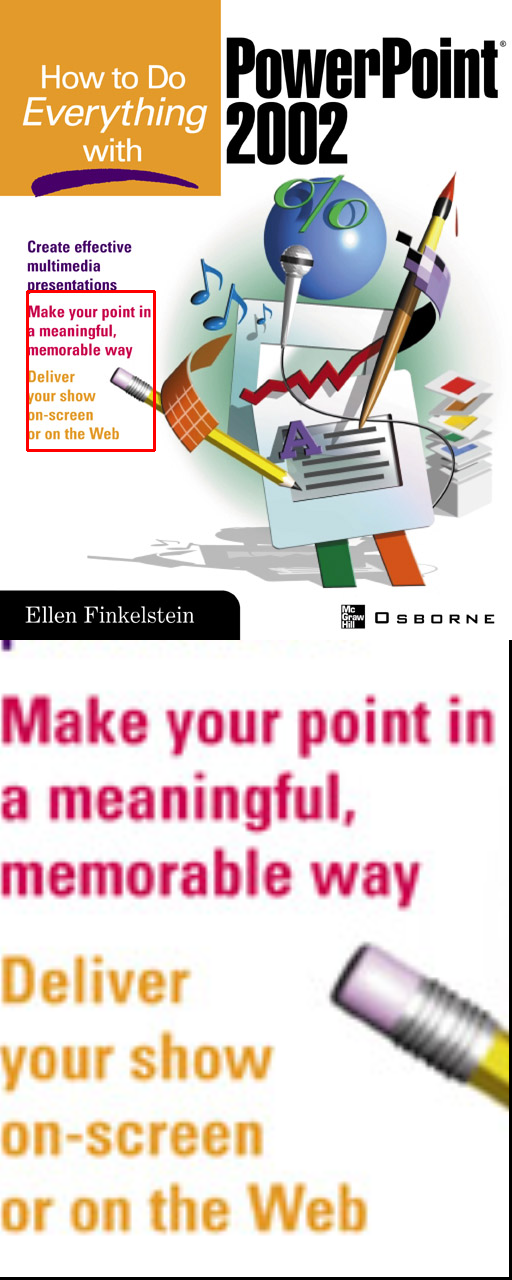}}
		\centerline{(a) ground truth image}\medskip
	\end{minipage}
	\begin{minipage}[b]{.19\linewidth}
		\centering
		\centerline{\includegraphics[width=\linewidth]{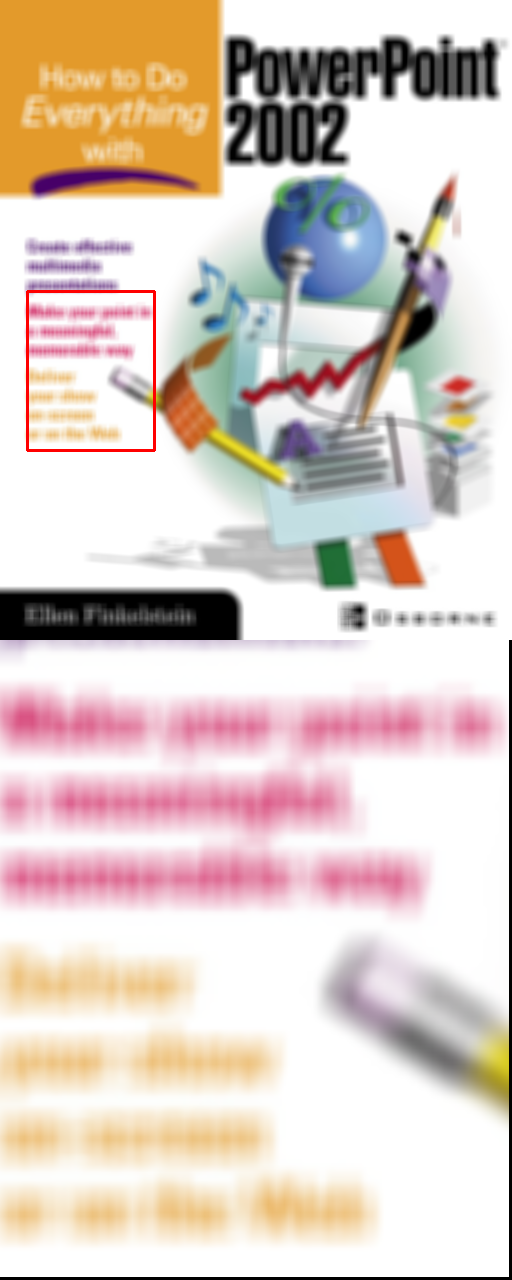}}
		\centerline{(b) blurred image}\medskip
	\end{minipage}
	\begin{minipage}[b]{.19\linewidth}
		\centering
		\centerline{\includegraphics[width=\linewidth]{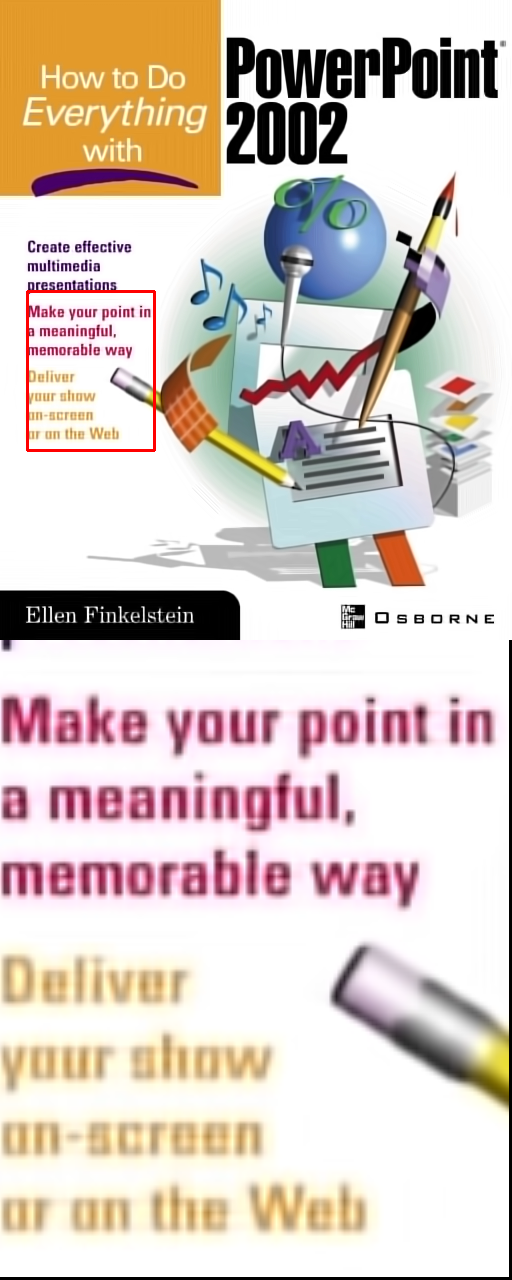}}
		\centerline{(c) NCSR (28.01dB)}\medskip
	\end{minipage}
        \begin{minipage}[b]{.19\linewidth}
		\centering
		\centerline{\includegraphics[width=\linewidth]{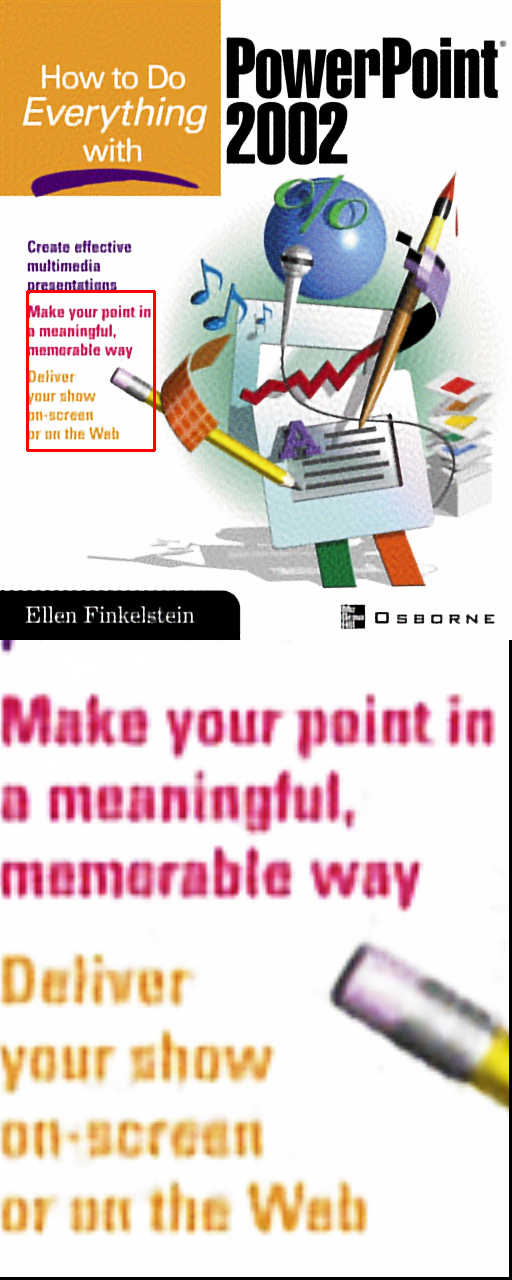}}
		\centerline{(d) DIP (29.49dB)}\medskip
	\end{minipage}
        \begin{minipage}[b]{.19\linewidth}
		\centering
		\centerline{\includegraphics[width=\linewidth]{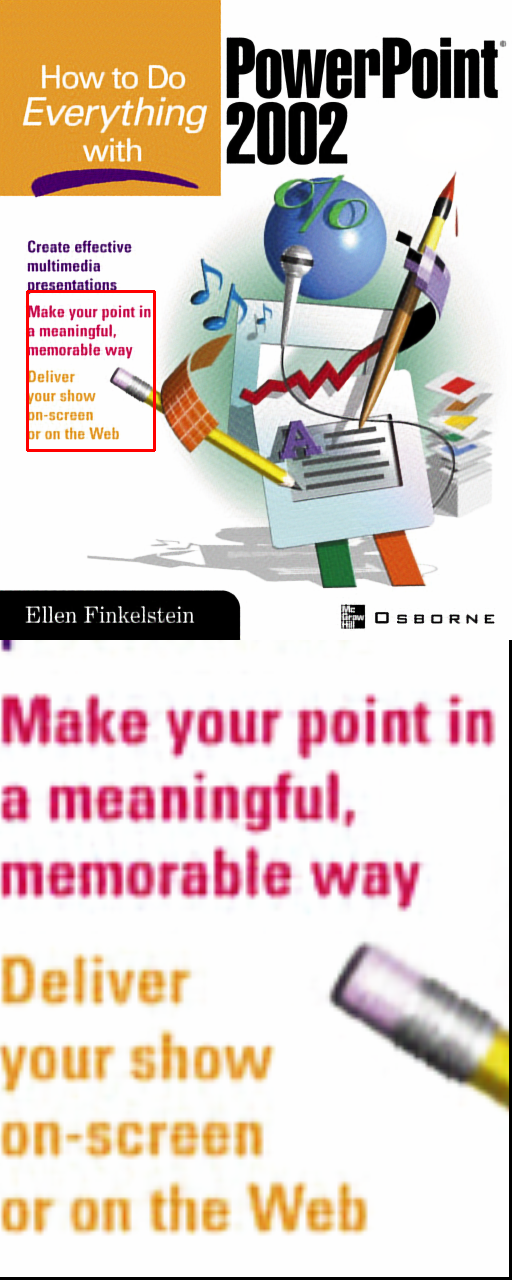}}
		\centerline{(e) SDIP (33.33dB)}\medskip
	\end{minipage}	
 \vspace{-0.4cm}
	\caption{Uniform deblurring result of PPT3.}
	\label{db1}
\end{figure*}

\begin{figure*}
	\centering
	\begin{minipage}[b]{.19\linewidth}
		\centering
		\centerline{\includegraphics[width=\linewidth]{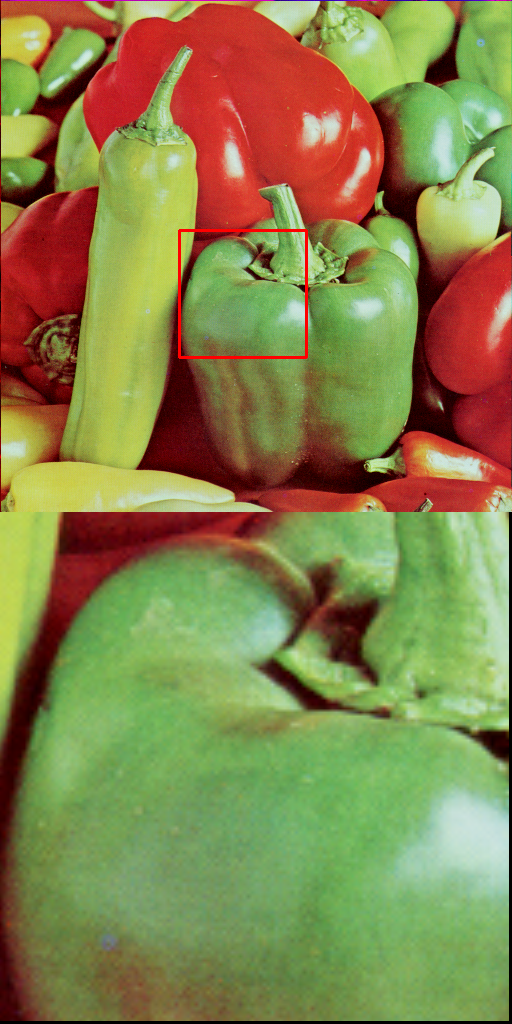}}
		\centerline{(a) ground truth image}\medskip
	\end{minipage}
	\begin{minipage}[b]{.19\linewidth}
		\centering
		\centerline{\includegraphics[width=\linewidth]{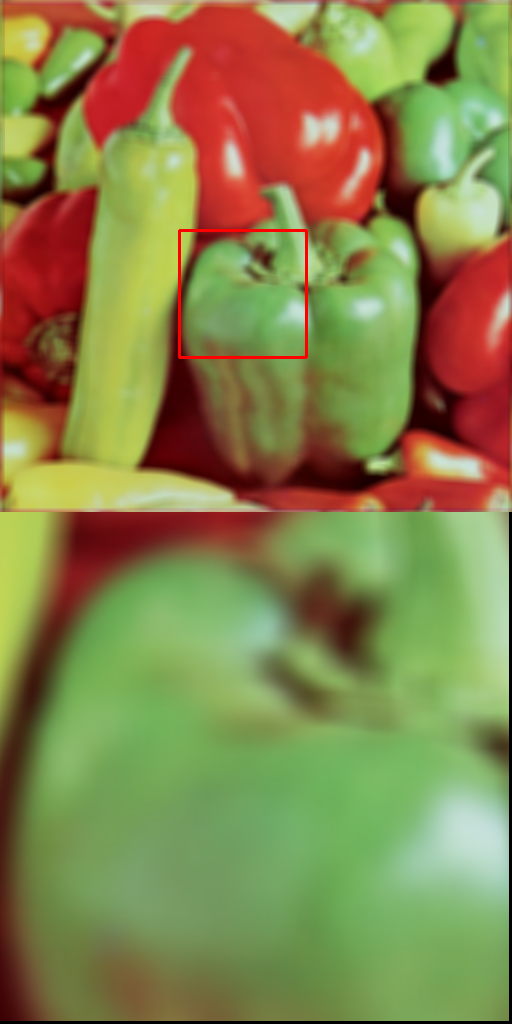}}
		\centerline{(b) blurred image}\medskip
	\end{minipage}
	\begin{minipage}[b]{.19\linewidth}
		\centering
		\centerline{\includegraphics[width=\linewidth]{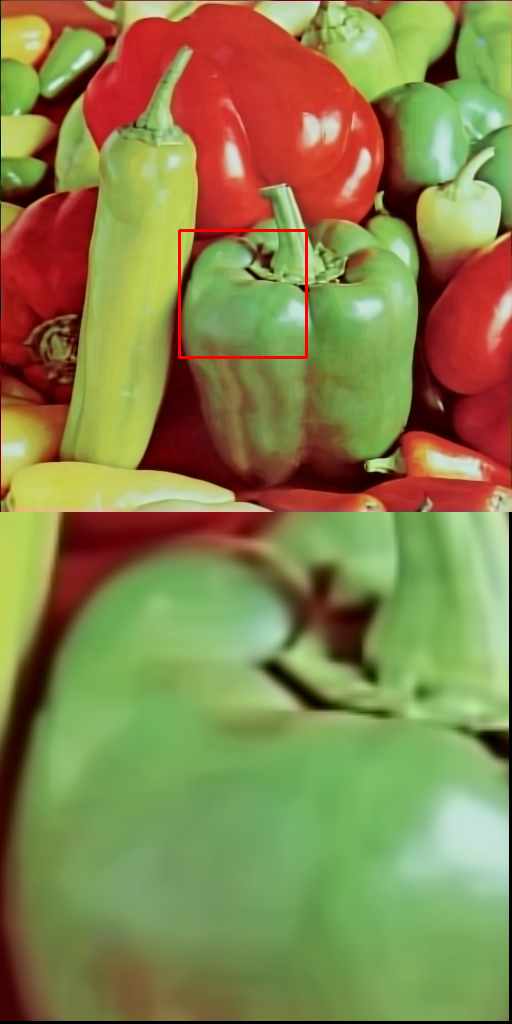}}
		\centerline{(c) NCSR (28.04dB)}\medskip
	\end{minipage}
        \begin{minipage}[b]{.19\linewidth}
		\centering
		\centerline{\includegraphics[width=\linewidth]{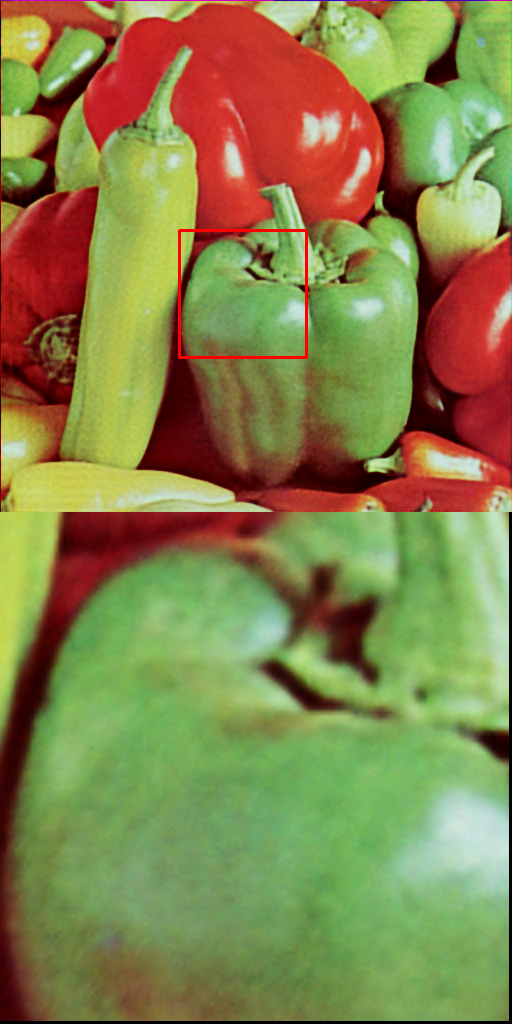}}
		\centerline{(d) DIP (29.96dB)}\medskip
	\end{minipage}
        \begin{minipage}[b]{.19\linewidth}
		\centering
		\centerline{\includegraphics[width=\linewidth]{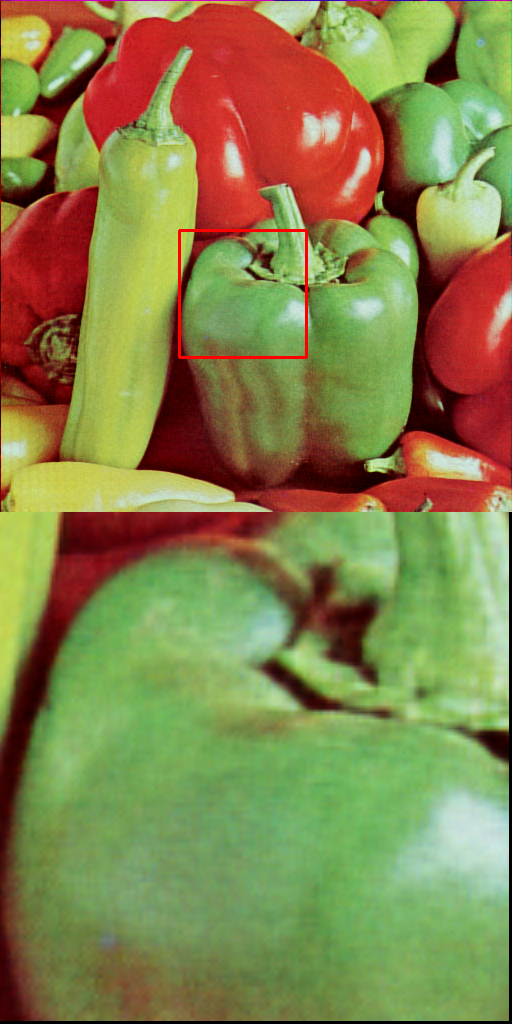}}
		\centerline{(e) SDIP (31.68dB)}\medskip
	\end{minipage}
 \vspace{-0.4cm}
	\caption{Uniform deblurring result of pepper.}
	\label{db2}
\end{figure*}

The experiments in this section are aligned with those described in~\cite{romano2017little} and~\cite{yair2018multi}, where the challenge involves a blurred image subject to a known degradation process. Our goal is to recover the original ground truth image from its blurred version. A uniform blur with a $9 \times 9$ kernel and a Gaussian blur characterized by a $25 \times 25$ kernel with a standard deviation ($\sigma$) of $1.6$ are conducted in our experiments. Our analysis focuses on comparing the improvement of the proposed SDIP algorithm over the original DIP method and the well-known NCSR deblur algorithm~\cite{dong2012nonlocally}. SDIP-GT is also conducted to further assess the efficacy of the self-reinforcement mechanism inherent to SDIP. These experiments are performed using images from the Set 5~\cite{bevilacqua2012low} and Set 14~\cite{zeyde2012single} datasets, the corresponding deblurring PSNR(dB) is shown in Table \ref{tab:db5} and Table \ref{tab:db14}. Fig.\ref{db1} and Fig.\ref{db2} illustrate the original ground truth images, their blurred counterparts, and the results generated by various deblurring algorithms. These results clearly demonstrate SDIP's great potential for image deblurring. Also, the superior performance of SDIP-GT further verifies the effectiveness of the self-reinforcement mechanism.

\begin{table}
  \centering
  \caption{The deblurring PSNR(dB) of different methods for the Set 5 dataset}
  \setlength{\tabcolsep}{0.5mm}{}
  \begin{tabular}{@{}lcccccc@{}}
  \toprule
  \multicolumn{7}{c}{Uniform Deblurring Results} \\
    \toprule
    Algorithm & baby & bird & butterfly & head & woman & average\\
    \midrule
    NCSR & 32.73 & 29.33 & 27.14 & 29.57 & 31.08 & 29.97\\
    DIP & 32.82 & 31.28 & 28.61 & 29.79 & 30.29 & 30.56\\
    SDIP-GT & 41.83 & 41.18 & 34.65 & 34.89 & 39.33 & 38.38\\
    SDIP & \textbf{34.89} & \textbf{32.82} & \textbf{28.80} & \textbf{31.10} & \textbf{31.61} & \textbf{31.84}\\
    \toprule
    \multicolumn{7}{c}{Gaussian Deblurring Results} \\
    \toprule
    Algorithm & baby & bird & butterfly & head & woman & average\\
    \midrule
    NCSR & 31.18 & 28.12 & 23.59 & 28.73 & 27.91 & 27.91\\
    DIP & 31.62 & 29.08 & 24.19 & 29.05 & 27.41 & 28.27\\
    SDIP-GT & 35.90 & 36.12 & 26.70 & 31.52 & 32.98 & 32.64\\
    SDIP & \textbf{32.74} & \textbf{31.26} & \textbf{24.93} & \textbf{29.71} & \textbf{28.04} & \textbf{29.33}\\
    \bottomrule
  \end{tabular}
  \label{tab:db5}
\end{table}

\begin{table*}
  \centering
  \caption{The deblurring PSNR(dB) of different methods for the Set 14 dataset}
  \setlength{\tabcolsep}{0.5mm}{}
  \begin{tabular}{@{}lccccccccccccc@{}}
  \toprule
  \multicolumn{14}{c}{Uniform Deblurring Results} \\
    \toprule
    Algorithm & baboon & barbara & coastgrd & comic & face & flowers & foreman & lenna & monarch & pepper & ppt3 & zebra & average\\
    \midrule
    NCSR & 22.00 & 26.27 & 29.12 & 23.73 & 29.57 & 25.99 & 32.35 & 30.55 & 28.05 & 28.04 & 28.01 & 28.92 & 27.72\\
    DIP & 21.97 & 24.96 & 29.10 & 23.93 & 29.38 & 26.79 & 30.64 & 30.54 & 27.71 & 29.96 & 29.49 & 28.23 & 27.73\\
    SDIP-GT & 30.64 & 38.81 & 41.43 & 34.87 & 33.94 & 37.21 & 38.29 & 36.17 & 36.94 & 36.23 & 39.79 & 40.61 & 37.08\\
    SDIP & \textbf{23.80} & \textbf{27.68} & \textbf{30.48} & \textbf{24.96} & \textbf{30.97} & \textbf{28.56} & \textbf{32.64} & \textbf{31.27} & \textbf{29.83} & \textbf{31.68} & \textbf{33.33} & \textbf{31.17} & \textbf{29.70}\\
    \toprule
    \multicolumn{14}{c}{Gaussian Deblurring Results} \\
    \toprule
    Algorithm & baboon & barbara & coastgrd & comic & face & flowers & foreman & lenna & monarch & pepper & ppt3 & zebra & average\\
    \midrule
    NCSR & 20.48 & 24.44 & 24.62 & 21.03 & 28.73 & 24.49 & 28.24 & 29.22 & 25.80 & 27.52 & 23.01 & 25.67 & 25.27\\ 
    DIP & 20.68 & 24.02 & 23.51 & 21.42 & 28.62 & 25.13 & 28.64 & 28.15 & 25.76 & 29.30 & 24.65 & 26.21 & 25.51\\ 
    SDIP-GT & 23.77 & 27.54 & 29.27 & 25.74 & 31.12 & 29.70 & 32.53 & 32.55 & 29.36 & 32.79 & 30.25 & 31.30 & 29.66\\
    SDIP & \textbf{21.15} & \textbf{24.55} & \textbf{25.55} & \textbf{21.87} & \textbf{29.70} & \textbf{25.52} & \textbf{29.92} & \textbf{30.04} & \textbf{24.51} & \textbf{30.04} & \textbf{25.31} & \textbf{26.89} & \textbf{26.26}\\ 
    \bottomrule
  \end{tabular}
  \label{tab:db14}
\end{table*}

\begin{figure*}
	\centering
	\begin{minipage}[b]{.19\linewidth}
		\centering
        \centerline{\includegraphics[width=\linewidth]{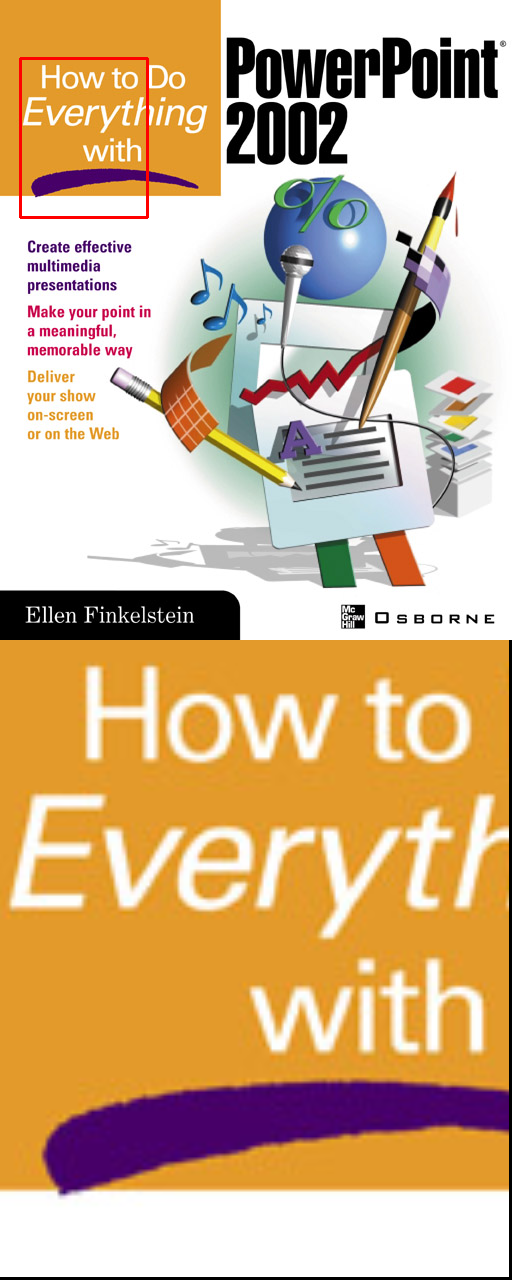}}
		\centerline{(a) ground truth image}\medskip
	\end{minipage}
	\begin{minipage}[b]{.19\linewidth}
		\centering
		\centerline{\includegraphics[width=\linewidth]{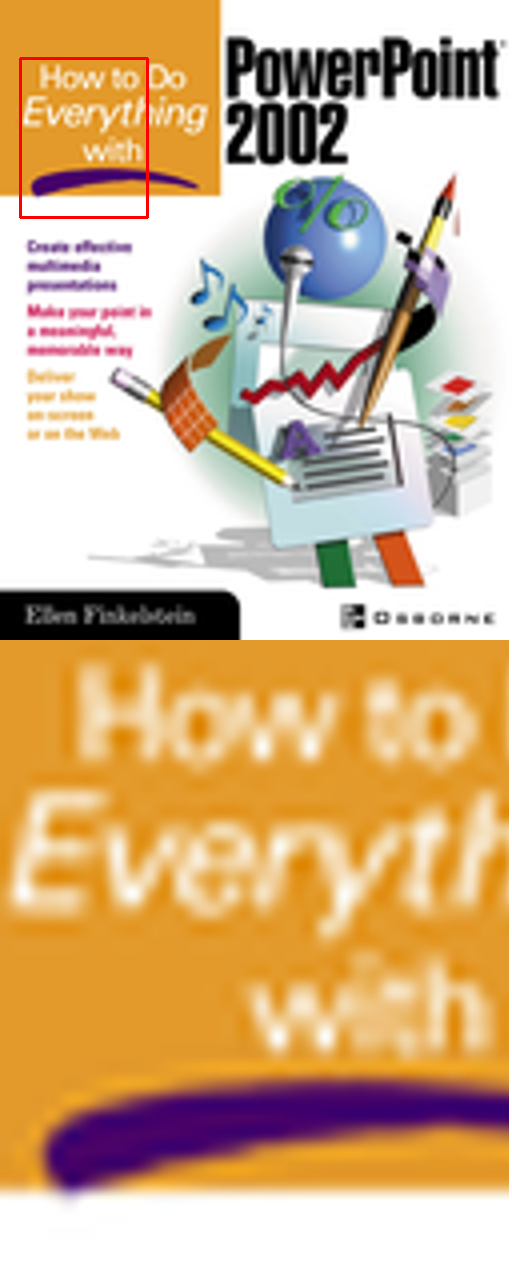}}
		\centerline{(b) bicubic (20.30dB)}\medskip
	\end{minipage}
	\begin{minipage}[b]{.19\linewidth}
		\centering
		\centerline{\includegraphics[width=\linewidth]{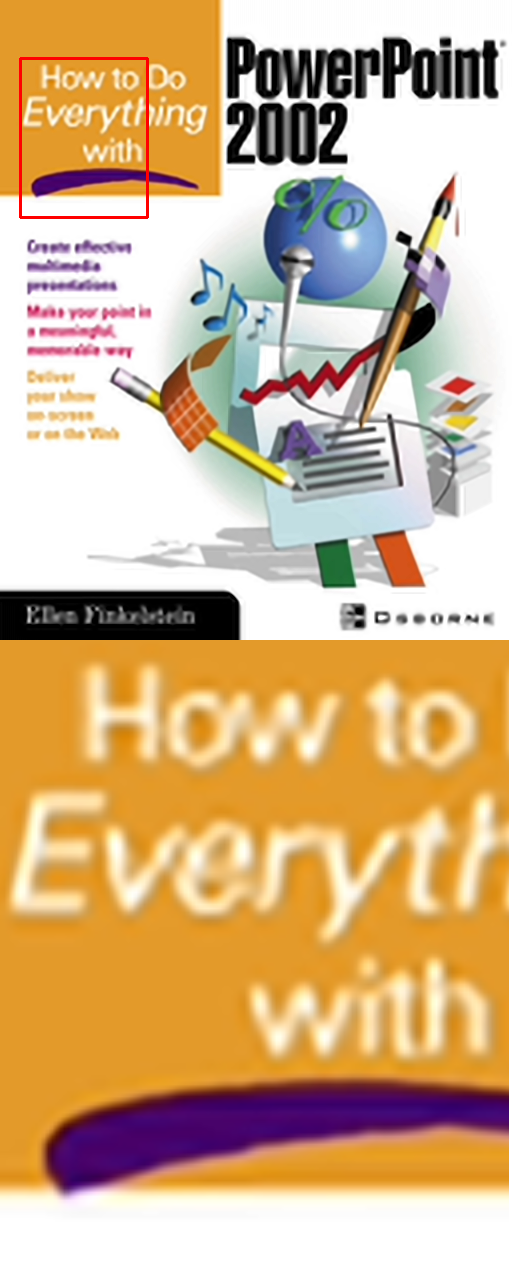}}
		\centerline{(c) NCSR (21.76dB)}\medskip
	\end{minipage}
        \begin{minipage}[b]{.19\linewidth}
		\centering
		\centerline{\includegraphics[width=\linewidth]{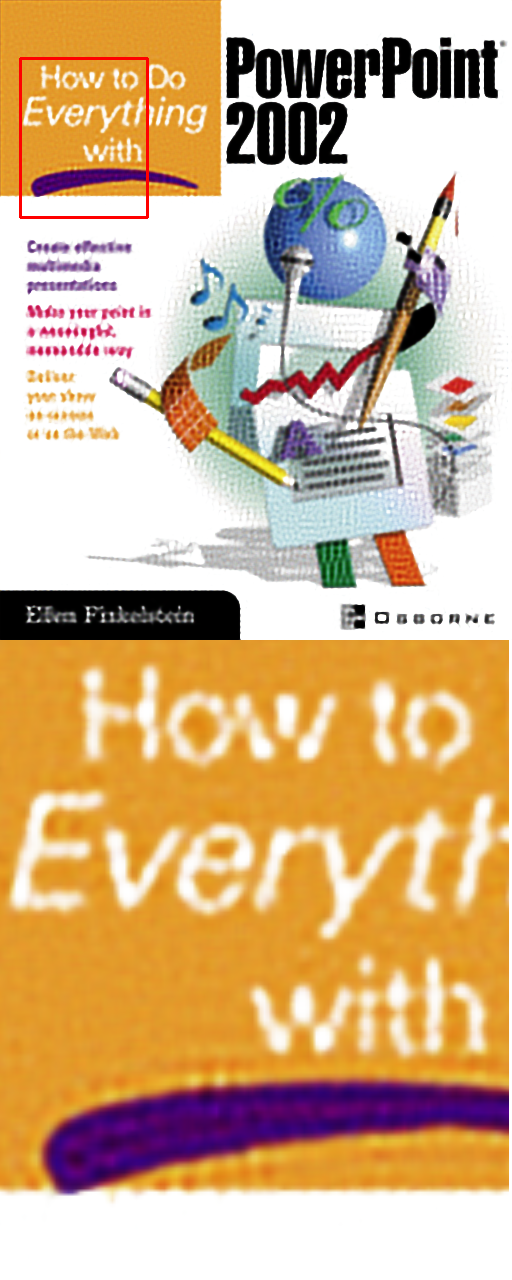}}
		\centerline{(d) DIP (23.93dB)}\medskip
	\end{minipage}
        \begin{minipage}[b]{.19\linewidth}
		\centering
		\centerline{\includegraphics[width=\linewidth]{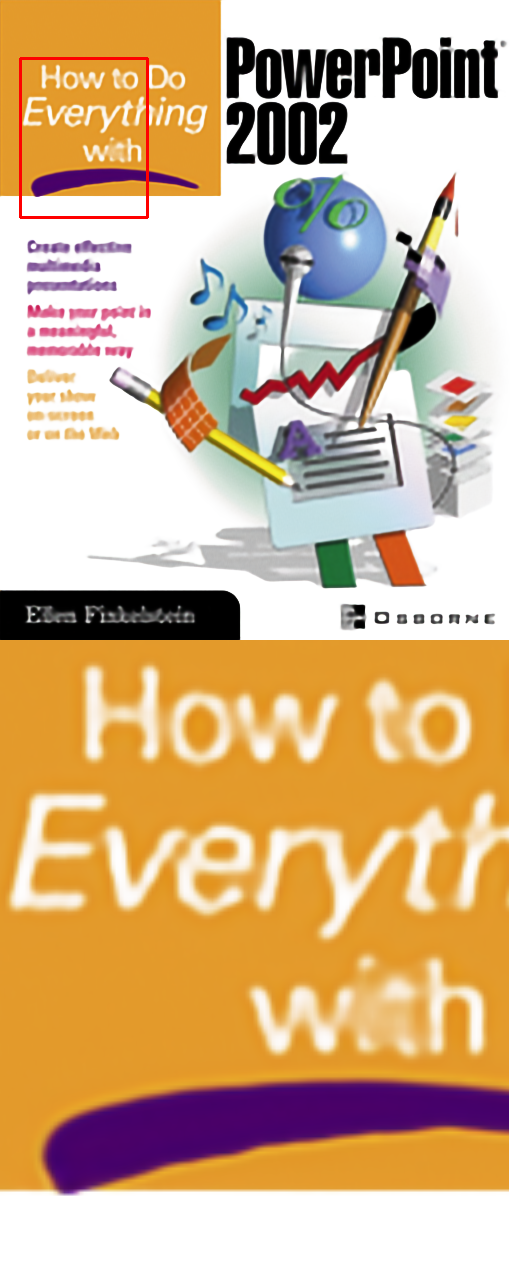}}
		\centerline{(e) SDIP (24.46dB)}\medskip
	\end{minipage}
 \vspace{-0.4cm}
	\caption{Super-resolution result of PPT3 with scale-factor 4.}
	\label{sp1}
\end{figure*}

\begin{figure*}
	\centering
	\begin{minipage}[b]{.19\linewidth}
		\centering
		\centerline{\includegraphics[width=\linewidth]{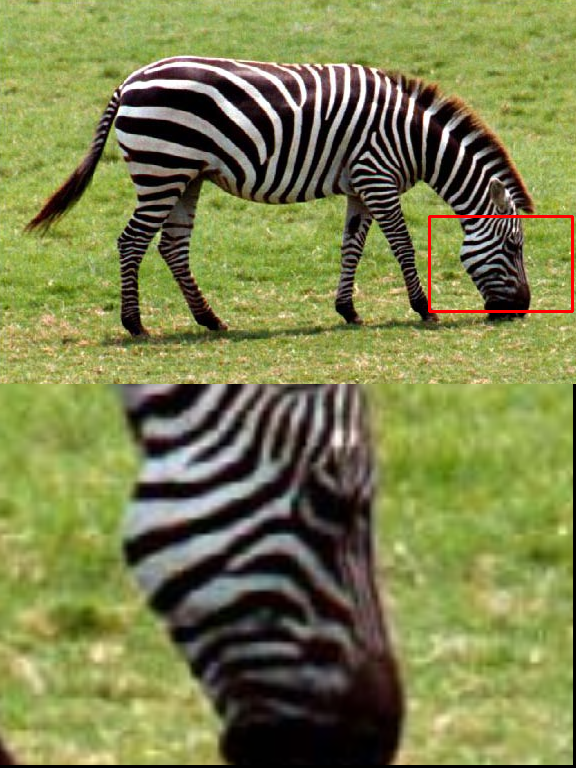}}
		\centerline{(a) ground truth image}\medskip
	\end{minipage}
	\begin{minipage}[b]{.19\linewidth}
		\centering
		\centerline{\includegraphics[width=\linewidth]{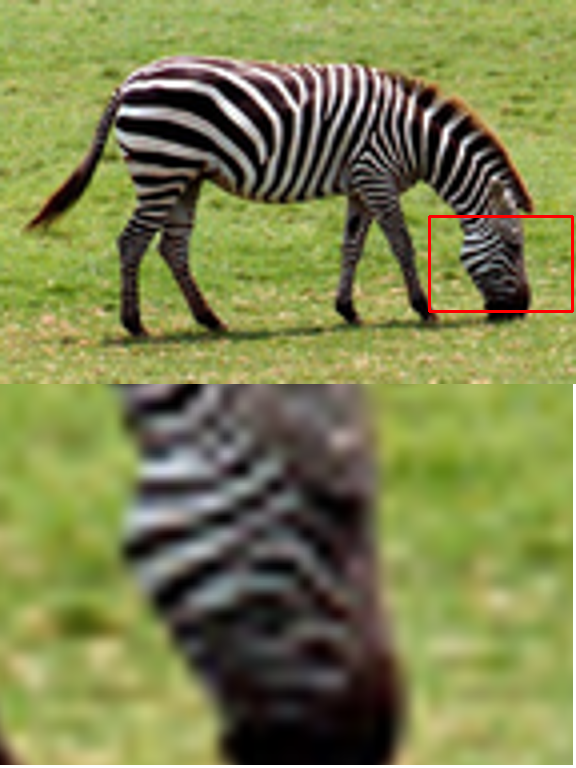}}
		\centerline{(b) bicubic (23.10dB)}\medskip
	\end{minipage}
	\begin{minipage}[b]{.19\linewidth}
		\centering
		\centerline{\includegraphics[width=\linewidth]{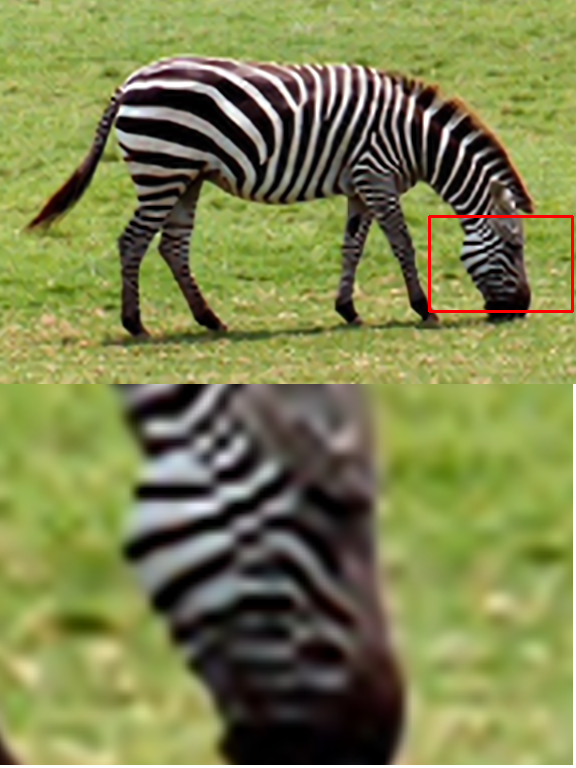}}
		\centerline{(c) NCSR (24.68dB)}\medskip
	\end{minipage}
        \begin{minipage}[b]{.19\linewidth}
		\centering
		\centerline{\includegraphics[width=\linewidth]{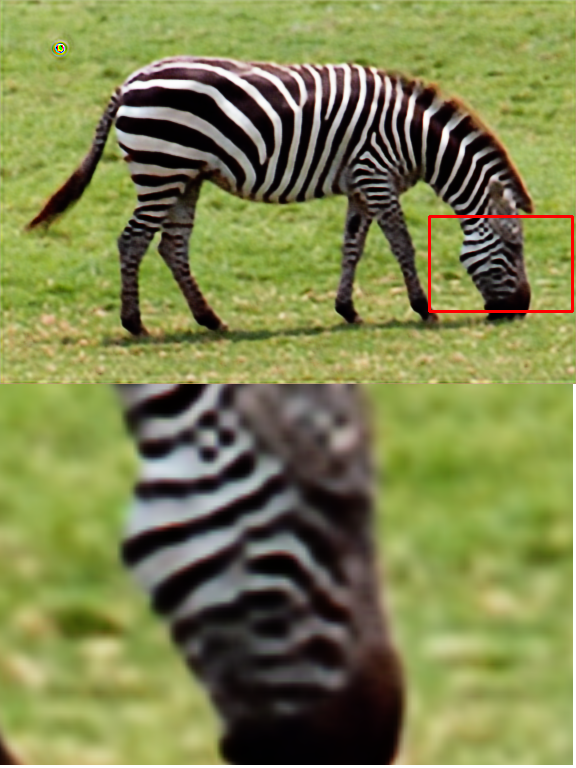}}
		\centerline{(d) DIP (23.40dB)}\medskip
	\end{minipage}
        \begin{minipage}[b]{.19\linewidth}
		\centering
		\centerline{\includegraphics[width=\linewidth]{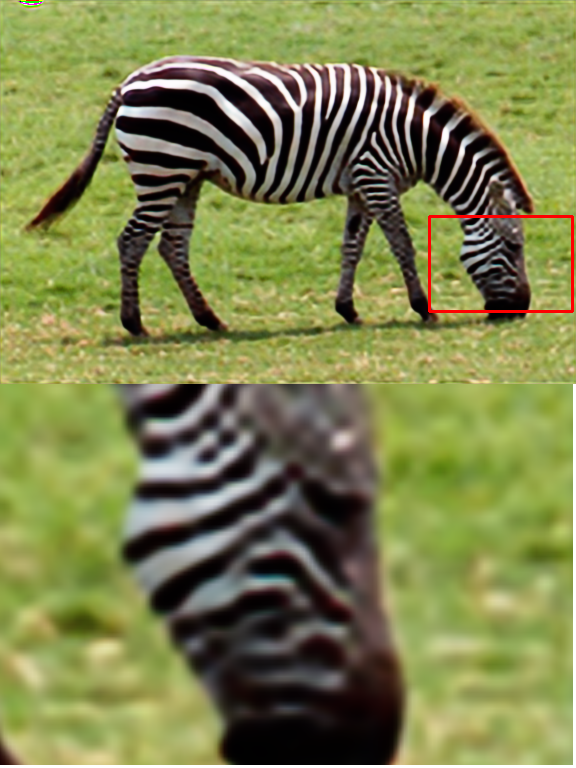}}
		\centerline{(e) SDIP (25.35dB)}\medskip
	\end{minipage}
    \vspace{-0.4cm}
	\caption{Super-resolution result of zebra with scale-factor 4.}
	\label{sp2}
\end{figure*}

\begin{table*}
  \centering
  \caption{The super-resolution PSNR(dB) of different methods for the Set 14 dataset}
  \setlength{\tabcolsep}{0.5mm}{}
  \begin{tabular}{@{}lccccccccccccc@{}}
    \toprule
    \multicolumn{14}{c}{Super Resolution x4} \\
    \toprule
    Algorithm & baboon & barbara & coastgrd & comic & face & flowers & foreman & lenna & monarch & pepper & ppt3 & zebra & average\\
    \midrule
    Bicubic & 20.30 & 23.57 & 24.05 & 20.20 & 28.82 & 23.76 & 26.11 & 28.56 & 24.55 & 27.34 & 20.30 & 23.10 & 24.22\\
    Nearest & 19.78 & 22.72 & 23.42 & 19.01 & 27.50 & 22.02 & 24.14 & 26.57 & 23.17 & 25.36 & 18.91 & 20.82 & 22.78\\
    NCSR & 20.40 & 24.11 & 24.61 & 20.98 & \textbf{29.45} & 24.57 & 27.84 & 29.15 & \textbf{25.64} & 27.46 & 21.76 & 24.68 & 25.05\\
    DIP & 19.74 & 22.62 & 23.24 & 19.64 & 28.35 & 23.46 & 26.94 & 25.50 & 24.79 & 27.92 & 23.93 & 23.40 & 24.13\\
    DIP-Gaussian & 20.15 & 23.58 & 23.91 & 20.64 & 28.80 & 24.87 & 28.18 & 27.98 & 25.10 & 25.23 & 23.98 & 23.39 & 24.65\\
    SDIP & \textbf{20.68} & \textbf{24.02} & \textbf{24.68} & \textbf{21.23} & 29.26 & \textbf{25.10} & \textbf{28.18} & \textbf{29.45} & 25.56 & \textbf{29.05} & \textbf{24.46} & \textbf{25.35} & \textbf{25.58}
\\
    \toprule
    \multicolumn{14}{c}{Super Resolution x8} \\
    \toprule
    Algorithm & baboon & barbara & coastgrd & comic & face & flowers & foreman & lenna & monarch & pepper & ppt3 & zebra & average\\
    \midrule
    Bicubic & 19.12 & 21.79 & 21.97 & 17.77 & 26.66 & 20.51 & 22.26 & 25.31 & 21.87 & 24.34 & 17.17 & 18.54 & 21.44 \\
Nearest & 18.63 & 20.83 & 21.38 & 16.91 & 25.11 & 19.27 & 20.74 & 23.63 & 20.69 & 22.38 & 16.27 & 17.69 & 20.29 \\
NCSR & 18.50 & 21.17 & 20.66 & 16.97 & 26.33 & 19.63 & 21.02 & 24.40 & 21.43 & 22.41 & 16.33 & 17.56 & 20.53\\
DIP & 17.88 & 20.64 & 21.23 & 17.40 & 25.10 & 20.39 & 22.85 & 23.86 & 21.91 & 23.05 & 17.88 & 18.48 & 20.89 \\
DIP-Gaussian & 19.05 & \textbf{22.07} & 22.31 & 18.20 & 26.46 & 21.32 & 24.62 & \textbf{26.22} & 22.41 & 25.97 & 19.25 & 18.87 & 22.23 \\
SDIP & \textbf{19.63} & 21.94 & \textbf{22.49} & \textbf{18.26} & \textbf{26.90} & \textbf{21.49} & \textbf{25.55} & 26.15 & \textbf{22.59} & \textbf{26.45} & \textbf{19.40} & \textbf{19.63} & \textbf{22.54} \\
    \bottomrule
  \end{tabular}
  \label{tab:s14}
\end{table*}

\subsection{Super-Resolution}

In this section, we evaluate the SDIP framework's efficacy in single-image super-resolution (SISR). Our objective is to recover the high-resolution image from its low-resolution counterpart. We test scaling factors 4 and 8, comparing them with methods such as the original DIP, NCSR super-resolution, bicubic interpolation, and nearest neighbor interpolation. It's worth mentioning that the authors of the DIP algorithm did not use the original DIP approach for SISR in their publication. In the original DIP algorithm, the network input is randomly initialized and then fixed in the later iterations. However, the author proposed using a random Gaussian vector to modify the input vector in each iteration when doing SISR, which we designate as DIP-Gaussian in our experiments. Interestingly, the DIP-Gaussian can be interpreted as modifying the network input with a random Gaussian vector iteratively to steer the network to improved results. This is similar to the methodology we proposed in the SDIP framework. To ensure a complete comparison, we incorporate both the DIP and DIP-Gaussian in our experiments on the Set 5 and the Set 14 datasets. These results are summarized in Table.\ref{tab:s5} and Table.\ref{tab:s14}. Fig.\ref{sp1} and Fig.\ref{sp2} present two visual results taken from these experiments to illustrate the recovery obtained by different methods. 

The results of our experiment reveal that DIP-Gaussian outperforms the original DIP, and SDIP exhibits even superior performance over DIP-Gaussian in most cases. This substantiates the feasibility of optimizing network output by modifying the network's input. It is worth mentioning that such improvements are not as pronounced in the context of super-resolution tasks with a scale factor of $8$. This diminished effectiveness is caused by the excessively large scale factor (a single pixel in the low-resolution image represents 64 pixels of the corresponding high-resolution image), which impedes the steering algorithm employed in this paper (a simple gradient descent algorithm minimizing the data inconsistency) from effectively extracting information. The adoption of more advanced algorithms could potentially address this limitation.

\begin{table}[H]
  \centering
  \caption{The super-resolution PSNR(dB) of different methods for the Set 5 dataset}
  \setlength{\tabcolsep}{0.5mm}{}
  \begin{tabular}{@{}lcccccc@{}}
    \toprule
    \multicolumn{7}{c}{Super Resolution x4} \\
    \toprule
    Algorithm & baby & bird & butterfly & head & woman & average\\
    \midrule
    Bicubic & 30.73 & 28.43 & 21.17 & 28.82 & 25.51 & 26.93 \\
    Nearest & 27.75 & 25.22 & 18.84 & 27.50 & 23.08 & 24.48 \\
    NCSR & \textbf{31.13} & 28.62 & 23.82 & \textbf{29.50} & 26.74 & 27.96 \\
    DIP & 30.13 & 26.26 & 22.01 & 28.05 & 24.86 & 26.26 \\
    DIP-Gaussian & 29.16 & 28.95 & 25.04 & 28.77 & 26.93 & 27.77 \\
    SDIP & 31.01 & \textbf{30.08} & \textbf{25.17} & 29.35 & \textbf{27.43} & \textbf{28.61} \\
    \toprule
    \multicolumn{7}{c}{Super Resolution x8} \\
    \toprule
    Algorithm & baby & bird & butterfly & head & woman & average\\
    \midrule
    Bicubic & 26.14 & 23.30 & 16.84 & 26.66 & 21.64 & 22.92 \\
    Nearest & 24.03 & 21.48 & 15.67 & 25.11 & 19.85 & 21.23 \\
    NCSR & 25.05 & 21.43 & 16.32 & 26.33 & 20.66 & 21.95 \\
    DIP & 24.80 & 20.66 & 17.50 & 25.31 & 21.86 & 22.03 \\
    DIP-Gaussian & \textbf{26.64} & 24.63 & \textbf{19.46} & 26.59 & 23.07 & 24.08 \\
    SDIP & 26.33 & \textbf{24.81} & 19.12 & \textbf{26.98} & \textbf{23.28} & \textbf{24.10} \\
    \bottomrule
  \end{tabular}
  \label{tab:s5}
\end{table}

\section{Discussion}
\label{sec:discussion}

In the previous section, we show that the SDIP outperforms the original DIP methods across various applications. Fig.\ref{reconresult} illustrates that the SDIP can reconstruct artifact-free CT images under few-view and limited-angle conditions, which correspond to highly ill-posed inverse problems. The fact that the SDIP needs no training datasets is particularly noteworthy in the domain of medical imaging, where high-quality training sets are often limited. The experiments also show the great potential of the SDIP in applications such as deblurring and super-resolution. It can restore image details with remarkable clarity, such as the text in Fig.\ref{db1} and Fig.\ref{sp1}, and the surface texture of the peppers in Fig.\ref{db2}.

To further prove that these advancements stem from the self-reinforcement mechanism of SDIP, Fig.\ref{reconresultfewview} illustrates the detailed limited-angle CT reconstruction process of SDIP. From Fig.\ref{reconresultfewview}, it is evident that the self-reinforcement mechanism can effectively correct missing elements and errors in images generated by the network. Additionally, we introduced an SDIP-GT variant in our experiments for comparison. Unlike the standard SDIP, the SDIP-GT's steering algorithm leverages the ground truth image to guide the process. The exceptional performance achieved by SDIP-GT not only validates the efficacy of the self-reinforcement mechanism but also suggests that the approach outlined in this paper has significant potential for further improvement. The self-reinforcement mechanism can use images obtained from other algorithms, utilizing different priors to further enhance the performance of SDIP.

As an enhancement of the DIP method, the SDIP also inherits some of the limitations of its predecessor, including the sensitivity to noise. Fortunately, the SDIP algorithm follows a procedure similar to that of the DIP, which means that most existing techniques used with DIP to counteract noise are also applicable to the SDIP framework, offering a pathway to mitigate this inherited vulnerability. For example, incorporating total variation regularization into the loss function~\cite{liu2019image}, introducing additional priors through the ADMM algorithm~\cite{mataev2019deepred}, or optimizing the stopping criterion~\cite{jo2021rethinking} are all viable strategies for enhancing the SDIP's performance.

\section{Future Works}
\label{sec:future}
Further research is required to understand and improve the effectiveness of this approach: 
\begin{itemize}
\item The use of more sophisticated steering algorithms could potentially further improve the performance. For example, a pre-trained neural network can be used as the steering algorithm. In that case, it can use the information learned from training data to guide the DIP network. Furthermore, as the objective function of the SDIP is optimized on inference data, interference from the training dataset can be suppressed.

\item Investigating how the SDIP algorithm can be introduced into existing DIP frameworks such as~\cite{mataev2019deepred,jo2021rethinking} to leverage their strengths and mitigate their weaknesses

\item Given that DIP and its related methods fundamentally utilize the structure of networks as a prior, employing different network architectures for the same task can yield varied outcomes. A deeper analysis of this aspect is crucial for a comprehensive understanding of all DIP-related methodologies.
\end{itemize}

\section{Conclusions}
\label{sec:conclusions}

{Image recovery is a critical domain in the field of computer vision, enabling the restoration and enhancement of image data, including image reconstruction, super-resolution, etc. The Deep Image Prior (DIP) algorithm leverages the inherent architecture of convolutional neural networks as a form of prior knowledge, demonstrating significant potential across various image recovery tasks. However, the original DIP algorithm often lacks stability as the entire method is initialized randomly. In this paper, we propose the self-reinforcement deep image prior (SDIP) framework to further improve its performance. Our proposed SDIP framework integrates additional priors into the DIP framework by incorporating a steering algorithm and a self-reinforcement mechanism, thereby overcoming some of the challenges faced by the original DIP method. The experimental evaluations demonstrate that SDIP outperforms the standard DIP algorithm across multiple domains.

{\small
\bibliographystyle{ieee_fullname}
\bibliography{egbib}
}

\begin{IEEEbiography}
[{\includegraphics[width=1in,clip]{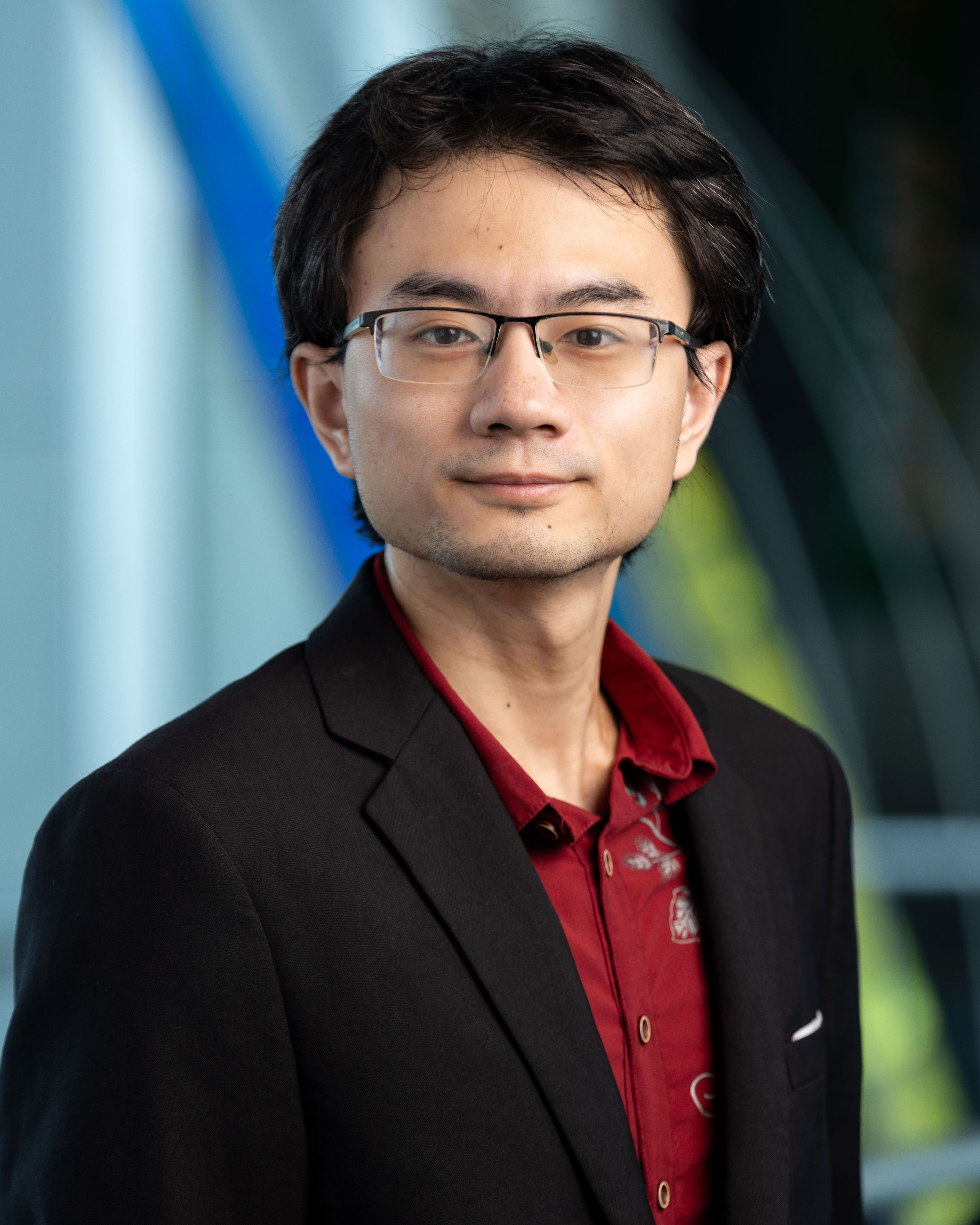}}]{Zhixin Pan} is an assistant professor in the Department of Electronic and Computer Engineering at the Florida State University. He received his Ph.D. in Computer Science from the University of Florida in 2022. His research interests include machine learning, cybersecurity, hardware security and trust, energy-aware computing, and quantum computing. He currently serves as an Assistant Editor of IEEE Transactions on VLSI Systems and ACM Transactions on Embedded Computing Systems. ea of research includes Cyber \& Hardware Security, post-silicon debugging, data mining, and machine learning.
\end{IEEEbiography}

\begin{IEEEbiography}
[{\includegraphics[width=1in,clip]{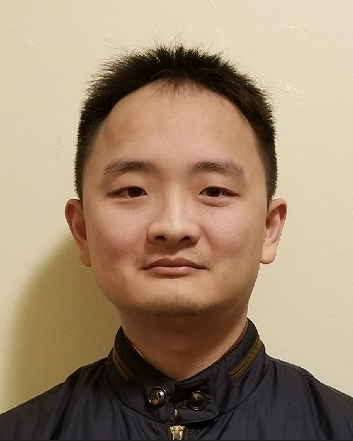}}]{Ziyu Shu} is a postdoctoral researcher in the Department of Radiation Oncology at the Washington University in St Louis. He received his Ph.D. in Computer Science from the University of Florida in 2023. His research interests include machine learning, medical imaging, and computer vision.
\end{IEEEbiography}
\end{document}